\documentclass[]{gtech}
\usepackage{amssymb}
\usepackage{multirow}
\usepackage{bigdelim}
\usepackage{todonotes}
\usepackage{longtable}
\usepackage{tabularray}
\usepackage{wrapfig}
\usepackage[most]{tcolorbox} 
\usepackage{xcolor}
\usepackage{url}
\usepackage{hyperref} 
\usepackage{float}
\usepackage[table,xcdraw]{xcolor}
\usepackage[normalem]{ulem}
\useunder{\uline}{\ul}{}

\renewcommand{\title}[1]{\newcommand{\titlelist}{{\huge\fontfamily{optimistic}\selectfont #1}}}

\newcommand{\model}{\texttt{DefakerOne}}
\newcommand{\modelbase}{\model}

\definecolor{prompt}{HTML}{5f84e4}
\definecolor{img}{HTML}{820100}

\usepackage{arydshln}

\definecolor{CQColor}{rgb}{0.0,0.0,1.0} 

\definecolor{TSColor}{rgb}{0.5,0.0,0.8} 

\definecolor{CQRColor}{rgb}{1.0,0.0,1.0} 

\usepackage{colortbl}
\usepackage{amssymb}
\usepackage{pifont}
\usepackage{booktabs,multirow}
\usepackage{makecell}
\usepackage{tabulary}
\usepackage{fontawesome5}
\usepackage{bbding}
\usepackage{multicol}

\newlength\savewidth

\title{
\raisebox{-0.55em}{\includegraphics[height=2.0em]{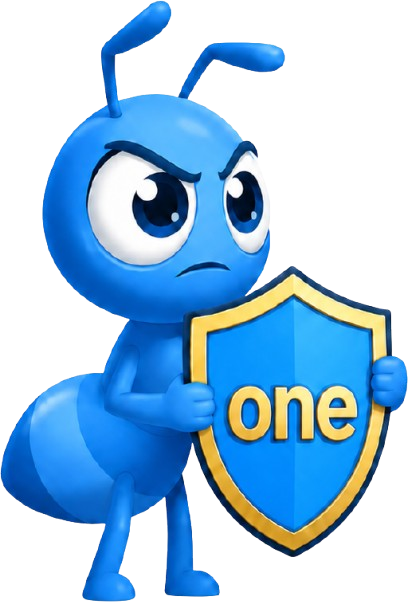}}\hspace{0.05em}
\textcolor[HTML]{0369ff}{Venus-DeFakerOne}: Uni{f}{i}ed Fake Image Detection \& Localization
}
\author{GuangJian Team, ~Ant Group}

\abstract{\fontsize{11pt}{12pt} \textit{In recent years, the rapid evolution of generative AI has fundamentally reshaped the paradigm of image forgery, breaking the traditional boundaries between document editing, natural image manipulation, DeepFake generation, and full-image AIGC synthesis. Despite this shift toward unified forgery generation, existing research in Fake Image Detection and Localization (FIDL) remains fragmented. 
This creates a mismatch between increasingly unified forgery generation mechanisms and the domain-specific detection paradigm. 
Bridging this mismatch poses two key challenges for FIDL: understanding cross-domain artifacts transfer and interference, and building a high-capacity unified foundation model for joint detection and localization.
To address these challenges, we propose DeFakerOne, a data-centric, unified FIDL foundation model integrating InternVL2 and SAM2. DeFakerOne enables simultaneous image-level detection and pixel-level forgery localization across diverse scenarios. Extensive experiments demonstrate that DeFakerOne achieves state-of-the-art performance, outperforming baselines on 39 forgery detection benchmarks and 9 localization benchmarks. Furthermore, the model exhibits superior robustness against real-world perturbations and state-of-the-art generators such as GPT-Image-2. Finally, we provide a systematic analysis of data scaling laws, cross-domain artifacts transfer-interference patterns, the necessity of fine-grained supervision, and the original resolution artifacts preservation, highlighting the design principles for scalable, robust, and unified FIDL.
}}

\date{May 15, 2026\vspace{-1mm}}

\gtechdata[Code]{%
    \faGithub \ \href{https://github.com/venus-guangjian/Venus-DeFakerOne}{\url{https://github.com/venus-guangjian/Venus-DeFakerOne}}%
}

\begin{document}
\maketitle

\begin{figure*}[h]
    \centering
    \includegraphics[width=0.98\textwidth]{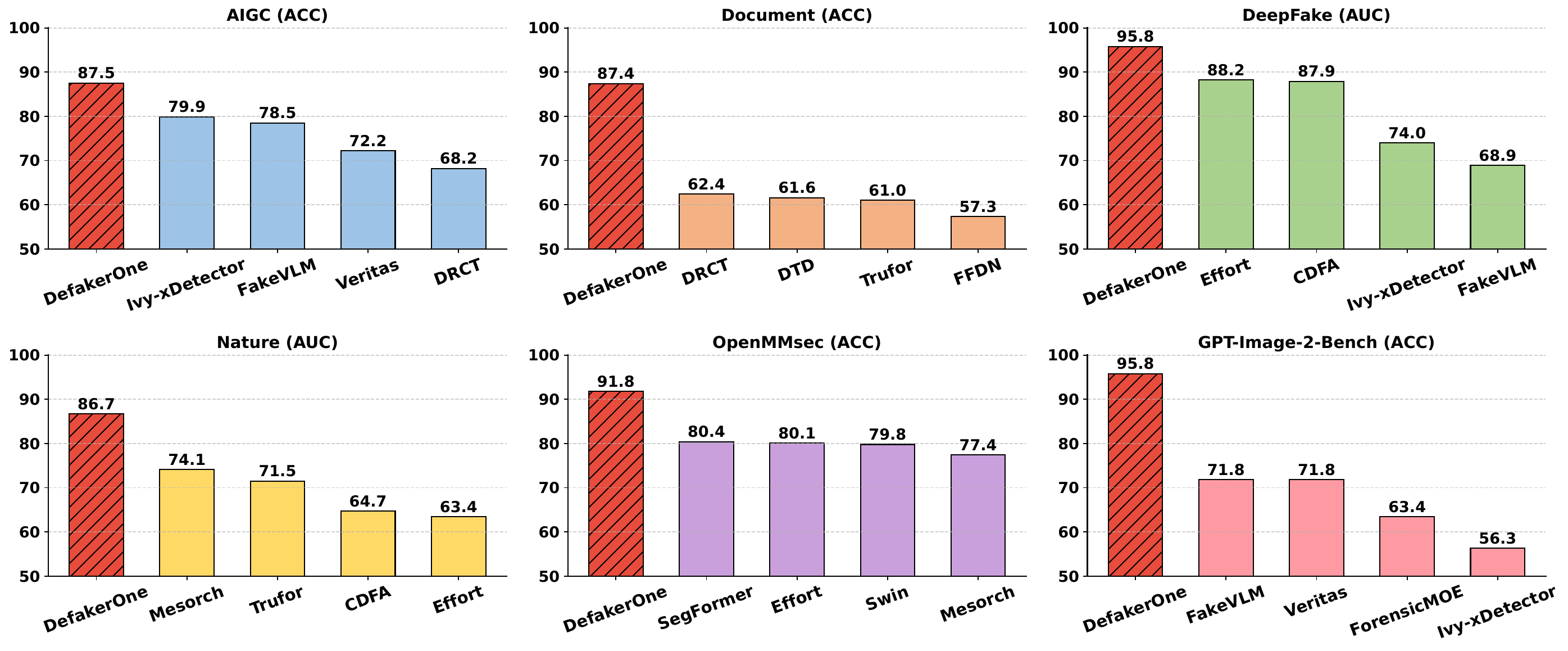}
    \caption{
    Overview of image-level forgery detection performance across six representative benchmarks. 
    DefakerOne achieves consistently superior performance on AIGC, DeepFake, document, natural-image, and GPT-Image-2 generated-image domains.
    }
    \label{fig:teaser}
\end{figure*}

\section{Introduction}
\label{sec:intro}

In recent years, the rapid development of AI-generated content has raised trust concerns in the digital world and brought increasing attention to digital forensics. Among related tasks, Fake Image Detection and Localization (FIDL)~\citep{forensichub} has become an important research direction.

\begin{figure}[t]
    \centering
    \includegraphics[width=0.95\linewidth]{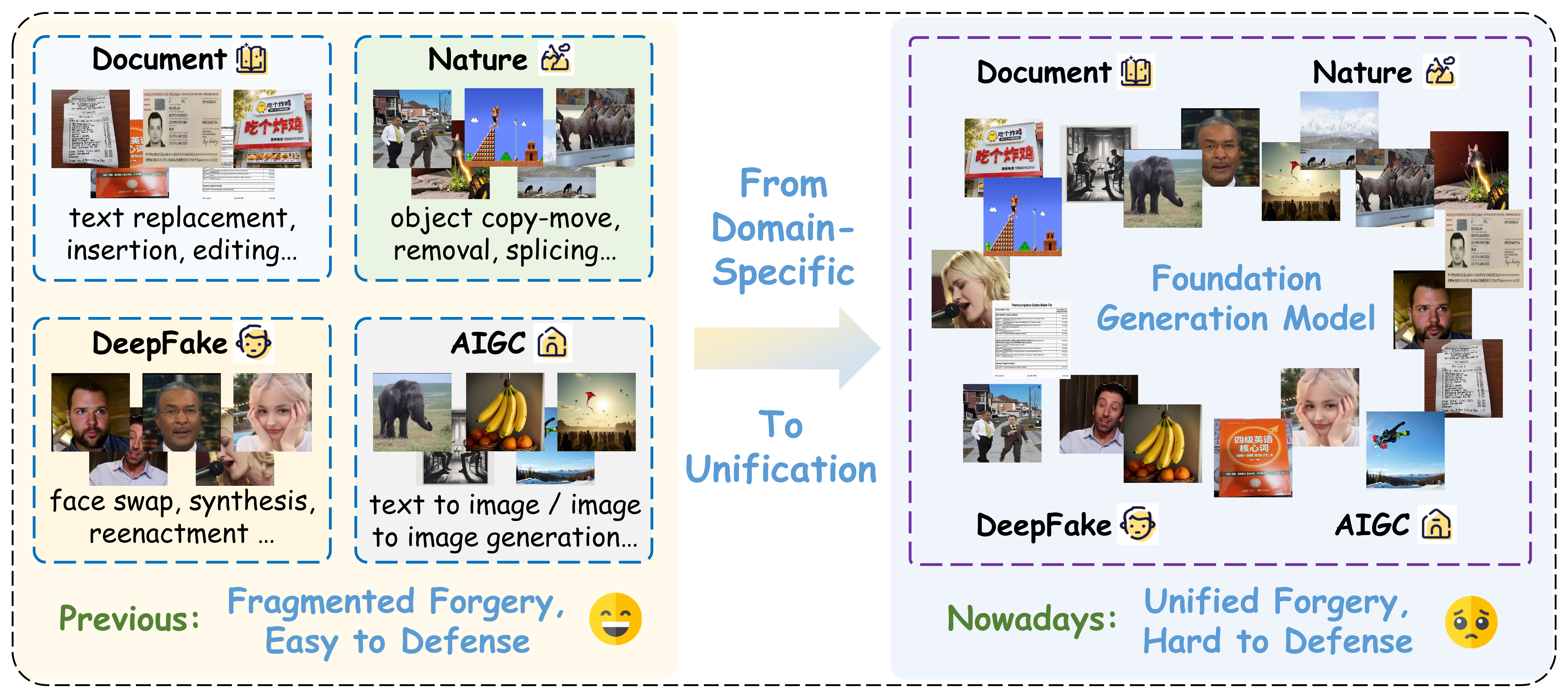}
\caption{
From domain-specific forgeries to unified forgery generation.
Earlier FIDL tasks are fragmented into Document, Nature, DeepFake, and AIGC, each relying on domain-specific manipulation operations and artifacts assumptions.
Foundation generation models now break the boundaries among these scenarios through shared generation and editing operations, making forgery artifacts more transferable and entangled across domains.
This evolution highlights the need for a unified FIDL model.
}
    \label{fig:intro_fidl_unification}
\end{figure}

Currently, as illustrated in Figure~\ref{fig:intro_fidl_unification}(a), due to the domain-specific characteristics of forged content and technology, FIDL research remains fragmented and is often divided into separate subfields, including document forgery detection~\citep{quomni, chen2024ffdn} (Document), natural image manipulation detection and localization~\citep{trufor2023, zhu2025mesorch} (Nature), face forgery detection~\citep{drct, dinomac} (DeepFake), and full-image AIGC detection~\citep{effort, univfd} (AIGC). These subfields usually adopt domain-specific methods. Document and Nature detection focus on local artifacts and semantic inconsistencies introduced by manual editing or glyph-guided generation methods such as AnyText~\citep{tuo2024anytext}. DeepFake detection focuses on local texture anomalies, such as moiré patterns and boundary artifacts, as well as physiological inconsistencies caused by post-generation blending in methods such as FaceSwap. AIGC detection mainly relies on global statistical distribution shifts and generation artifacts.

However, with recent advances in foundation generative models, the paradigm of image forgery is changing. As illustrated in Figure~\ref{fig:intro_fidl_unification} (b), powerful foundation generation models for text-to-image generation and image-to-image editing (T2I/I2I) are breaking the boundaries of previous scenarios. With open-domain generation and editing capabilities, they can cover forged documents, natural scenes, face images, social media content, and full-image synthesis. As a result, different forgery targets increasingly share similar mechanisms for generation, repainting, and editing, challenging previous single-domain detection assumptions based on specific objects, operations, or artifacts. For example, when facing advanced generators such as GPT-Image-2~\citep{openai_gpt_image_2_announcement}, forged content often shows higher texture fidelity, semantic consistency, and cross-domain diversity. In this case, traditional detectors can no longer rely on domain-specific artifacts for discrimination, leading to performance drops and limited robustness. Therefore, FIDL requires a unified research paradigm that can model these shared manipulation artifacts across domains, rather than treating Document, Nature, DeepFake, and AIGC as isolated forensic problems.

However, moving toward this unified FIDL paradigm still faces two core challenges:
\begin{enumerate}

    \item \textbf{The lack of systematic modeling of multi-domain artifacts interactions.}
    Although different FIDL subfields share similar input and output forms, existing studies show that their forensic supervision granularity and underlying artifacts are different. AIGC focuses more on global generation artifacts, DeepFake emphasizes face identity consistency and blending artifacts, Document relies on text, layout, and local artifacts, while Nature focuses on regional inconsistency and pixel-level localization. Thus, multi-domain data provides a basis for unified modeling, but the sources, scales, and forms of their artifacts differ. Whether these artifacts are transferable or conflicting across domains remains underexplored.

    \item \textbf{The limited exploration of large-capacity models for unified FIDL.}
    Existing FIDL methods are still mainly dominated by a small vision model paradigm. Most detectors are designed for individual subfields and optimized around specific artifacts, which limits their feature-space capacity for representing diverse forgery traces across AIGC, DeepFake, Document, and Nature. As a result, these models struggle to jointly handle image-level detection, pixel-level localization, and cross-domain generalization within a unified framework. Therefore, exploring large-capacity models with stronger visual-semantic representation and unified output interfaces is necessary for moving FIDL from fragmented research toward a unified foundation-model paradigm.
\end{enumerate}

To address these challenges, we propose \textbf{DeFakerOne}, a data-centric unified FIDL foundation model. On the dataset side, we curate 12.5M training samples for unified FIDL, covering multiple forensic domains including AIGC, DeepFake, Document, and Nature. The data sources include public open-source datasets, samples from closed-source generators, and private real-world scenario data. On the model side, the DeFakerOne integrates a unified InternVL2-2B~\citep{chen2024internvl} + SAM2~\citep{sam2} architecture, enabling both image-level detection and pixel-level localization. On the evaluation side, we conduct systematic experiments on 40 benchmarks across four domains, covering both detection and localization tasks, to analyze the performance, generalization ability, and data composition patterns of a unified model in multi-domain FIDL scenarios. In addition, to evaluate the model's adaptability to recent closed-source generators, we further construct \textbf{GPT-Image-2-Bench}, which contains 71 test samples and covers real-world application scenarios such as documents, face images, natural scenes, AIGC images, posters, and social media content.

Experimental results demonstrate that DeFakerOne establishes a strong baseline for unified FIDL. 
At the domain level, DeFakerOne achieves the best average performance across all four major FIDL domains, including 95.8 AUC on DeepFake, 87.5 ACC on AIGC, 87.4 ACC on Document, and 86.7 AUC on Nature. 
At the benchmark level, DeFakerOne achieves state-of-the-art results on 39 forgery detection benchmarks and 9 localization benchmarks. 
Moreover, DeFakerOne achieves state-of-the-art performance on OpenMMsec~\citep{du2026SICA} and GPT-Image-2-Bench, further validating its generalization under both cross-domain evaluation and forgery detection scenarios involving more powerful generative foundation models.

\section{Data Construction}

\subsection{Training Data}

\subsubsection{Overall}
\begin{figure}
    \centering
    \includegraphics[width=\linewidth]{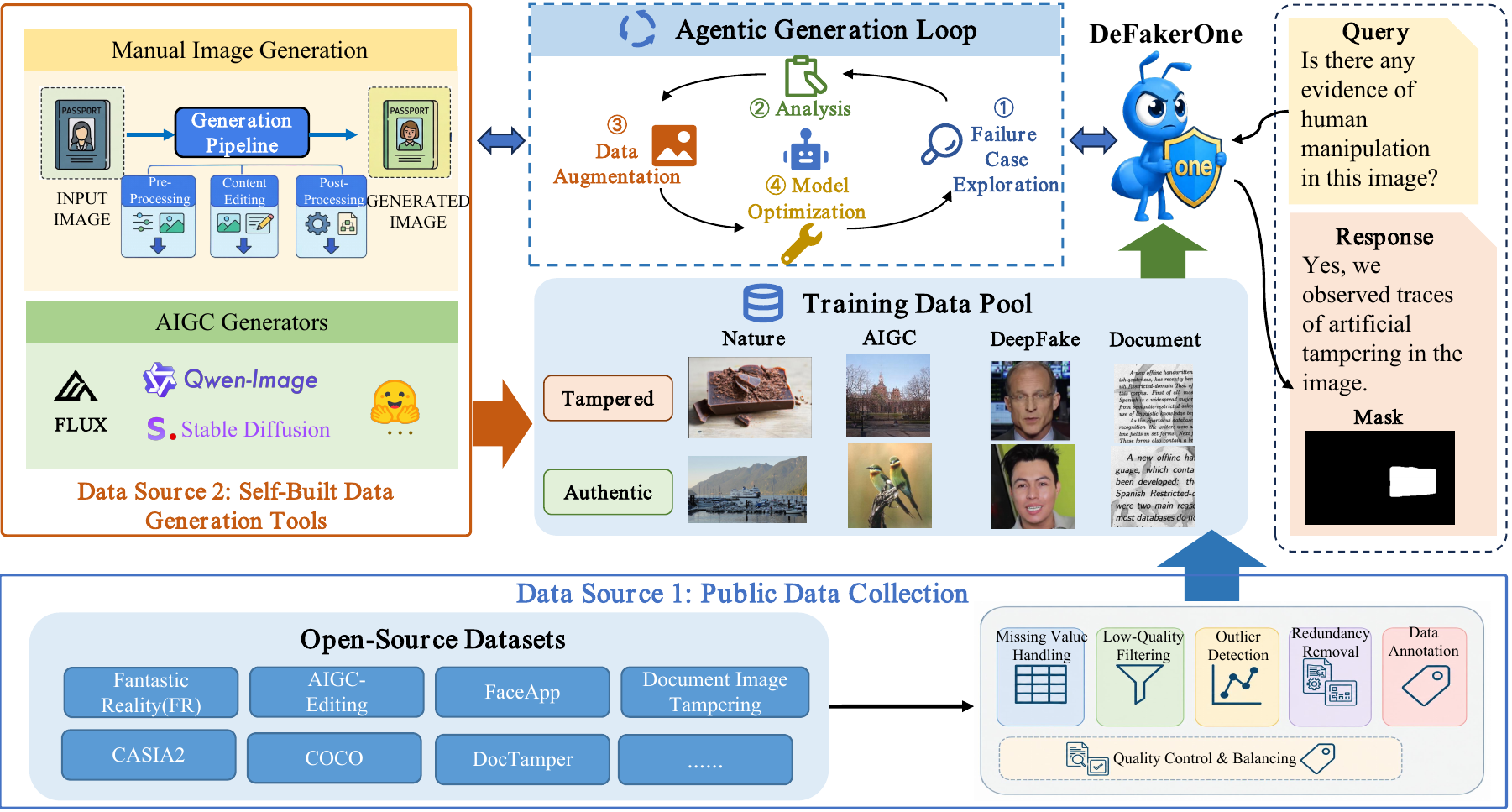}
    \caption{Closed-loop training based on difficult case mining. DeFakerOne is trained using data across four domains. Expert review of failed cases triggers reverse engineering of the manipulation chain, which in turn drives our generative agent to synthesize targeted augmented samples for iterative model optimization.}
    \label{fig:placeholder}
\end{figure}

We build our unified forgery detection model on a comprehensive, large-scale dataset that is meticulously curated to cover a wide spectrum of digital forgery techniques and content modalities. As illustrated in Figure \ref{fig:placeholder}, we built a diverse training dataset by aggregating multi-source data and establishing several dedicated data production pipelines. Our data strategy is staged, with each phase designed to achieve a specific training objective. 

\begin{figure}
    
    \centering
    \begin{subfigure}[c]{0.48\textwidth}
        \centering
        \includegraphics[width=\linewidth]{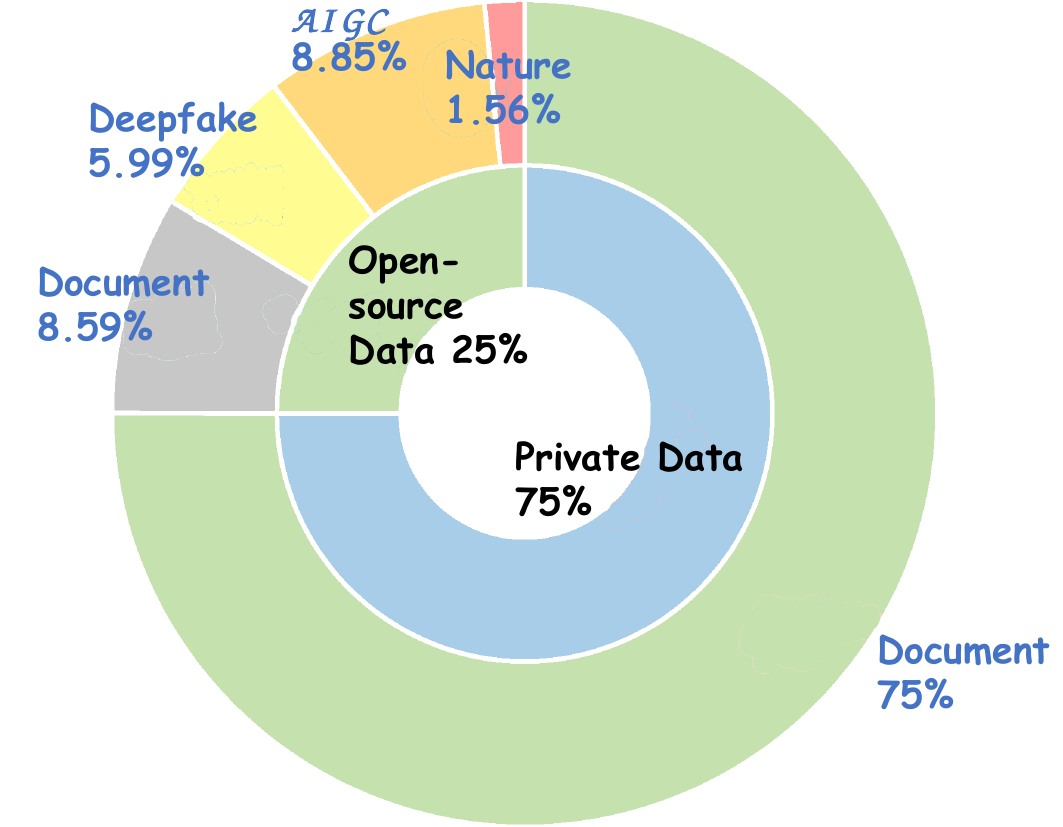}
        \caption{SFT-Stage-1-Data Image-Text:2M}
        \label{fig:stage1}
    \end{subfigure}
    \hfill
    \begin{subfigure}[c]{0.45\textwidth}
        \centering
        \includegraphics[width=\linewidth]{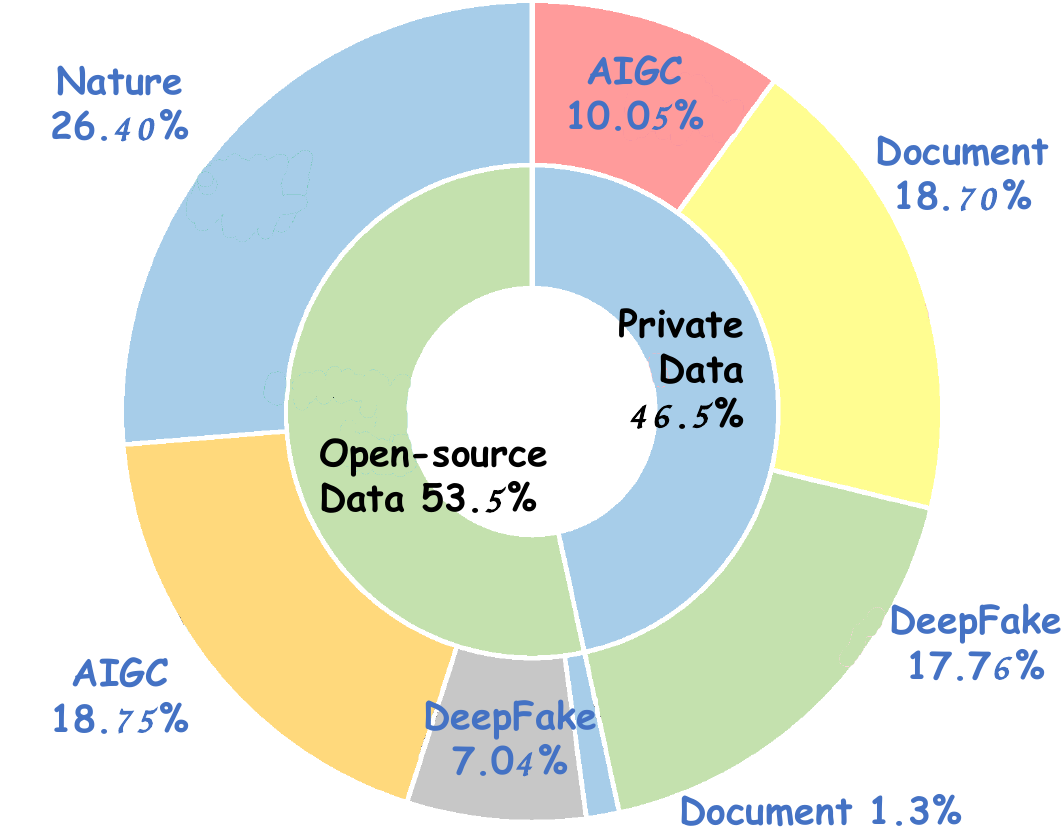}
        \caption{SFT-Stage-2-Data Image-Text:12.5M}
        \label{fig:stage2}
    \end{subfigure}
    
    \caption{
Data composition of the two-stage SFT training. 
(a) Stage-1 uses 2M image-text samples dominated by private document data. 
(b) Stage-2 expands to 12.5M samples with a more balanced mixture of open-source and private data across Document, DeepFake, AIGC, and Nature.
}
    \label{fig:data_composite}
\end{figure}

As shown in Figure~\ref{fig:data_composite}, our training data strategy adopts a staged design: Stage 1 focuses on private document-dominated image-text alignment, while Stage 2 expands to a balanced multi-domain mixture for unified FIDL training.

\textbf{Stage 1: Paradigm Validation}. We constructed an initial dataset of 2M samples, predominantly composed of private business data (75\%), supplemented by API-generated data and open-source benchmarks (25\%). This dataset covers AIGC, DeepFake, document, and natural image forensics domains. Through training in this stage, we validated that the model converges effectively across multiple tasks, achieving robust performance on both real-world scenarios and open-source data.

\textbf{Stage 2: Capability Expansion and Scaling}. We then scaled the dataset to 12.5M samples through five complementary pipelines: open-source datasets, private business data, API-based generation, high-quality expert PSD synthesis, and internal red-team adversarial samples. This three-tier data architecture—authentic, synthetic, and adversarial—drives our data scaling law experiments and extends model robustness against evolving forgery techniques.

\subsubsection{Data Pipeline}
To support data-centric training, we build a closed-loop data generation pipeline, as shown in Figure~\ref{fig:placeholder}. After DeFakerOne is trained on the data pool, its failure cases are collected and sent to an agent-assisted refinement module. The agent analyzes these bad cases to identify missing forgery patterns, difficult manipulation types, and domains, and then selects suitable generation or editing models to synthesize targeted samples. The newly generated data are added back to the training pool for the next round of optimization. Through this loop of training, agent-based bad-case analysis, data synthesis, and re-training, DeFakerOne can continuously adapt to evolving forgery methods.

Although the same agent-assisted refinement framework is applied across all FIDL domains, its execution process differs according to the characteristics of each domain. We describe these domain-specific pipeline designs below.

\textbf{For the AIGC domain}, we construct a large-scale dataset with 3.6M samples, including 2.344M public samples from DiffusionForensics~\citep{dire}, CommunityForensics~\citep{CommunityForensics}, GenImage~\citep{zhu2023genimage}, LAION\_DATA~\citep{schuhmann2022laion}, and ForenSynths~\citep{cnnspot}, together with 1.256M private curated samples.
In addition, we maintain a broad collection of commercial APIs and open-source generation/editing models.
When a new AIGC model appears, the agent automatically invokes the corresponding model or API, selects prompts and source images, and synthesizes new AIGC samples in batches.

\textbf{For the Document domain}, we collect 0.162M public samples from benchmarks such as DocTamper~\citep{qu2023towards}, T-SROIE~\citep{wang2022tsroie}, RTM~\citep{luo2025rtm}, SACP~\citep{sacp}, RIFLC~\citep{rific}, and OSTF~\citep{qu2025revisiting}, and further supplement them with 2.338M private real-world document samples.
We also built a private document pool covering more than 4,000 classes of real-world documents, credentials, contracts, invoices, and certificates.
For synthesis, the agent first matches the target document class and then applies suitable operations such as text replacement, seal modification, layout editing, and local region manipulation.

\textbf{For the DeepFake domain}, we collect 0.88M public samples from FaceForensics++~\citep{rossler2019faceforensics++}, CelebDF-v2~\citep{li2020celeb}, DFD~\cite{dfd2019}, DFDC~\cite{dfd2019}, ScaleDF~\citep{scaledf}, DF40~\citep{yan2024df40}, WDF~\citep{zi2020wilddeepfake}, and MFFI~\citep{miao2025mffi}, together with 2.22M private DeepFake samples.
We further prepare a large-scale open-source real-face pool covering diverse identities, head poses, facial expressions, lighting conditions, occlusions, backgrounds, and resolutions.
Similar to the AIGC pipeline, the agent invokes face manipulation models to synthesize DeepFake samples, while explicitly considering variations in face pose, expression, identity, and capture environment.

\textbf{For the Nature domain}, we curate 3.3M public natural-image samples from MIML~\citep{qu2024towards}, CASIA-v2~\citep{CASIA_2013}, COCO\_2017~\citep{mscoco_2014}, OpenSDI~\citep{opensdi}, So-Fake-OOD~\citep{huang2025sofakebenchmarkingexplainingsocial}, and So-Fake-Set~\citep{huang2025sofakebenchmarkingexplainingsocial}.
For synthesis, the agent uses pre-segmented regions and operation-specific prompts to generate local manipulations, including splicing, copy-move, object removal, inpainting, and generative local editing, with corresponding masks added to the training pool.

\begin{figure*}[t!]
    \centering
    \includegraphics[width=1.0\linewidth]{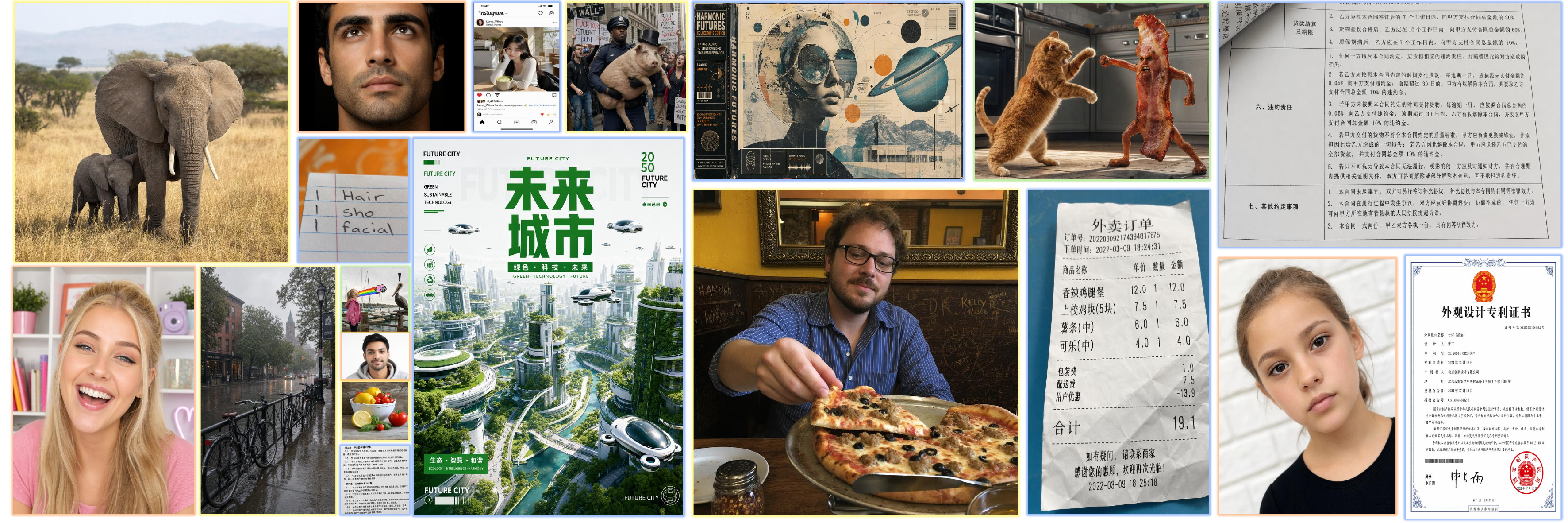}
    \caption{Some of the representative samples from the proposed GPT-Image-2-Bench.}
    \label{fig:vizgptbench}
\end{figure*}

\begin{figure}[t]
    \centering
    \includegraphics[width=0.68\textwidth]{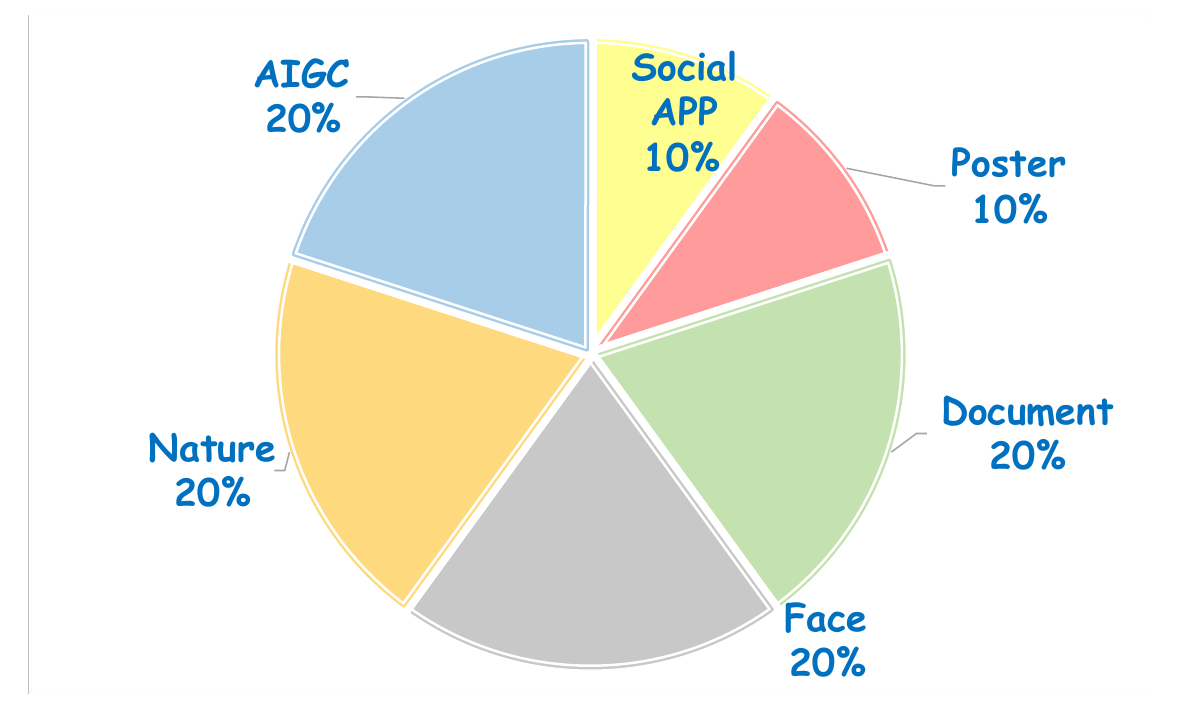}
    \caption{Data distribution of GPT-Image-2-Bench across six representative visual domains.}
    \label{fig:gpt-image-2-bench}
\end{figure}

\subsection{GPT-Image-2-Bench}
\label{sec:gpt_image_2_bench}

To evaluate the detection robustness against the latest generation foundation model~\citep{openai_gpt_image_2_announcement} generator, we construct GPT-Image-2-Bench, a benchmark containing 71 samples spanning diverse real-world scenarios. As shown in Figure~\ref{fig:gpt-image-2-bench}, the dataset is designed with a relatively balanced distribution: document (20\%), DeepFake (20\%), natural scenes (20\%), and general AIGC content (20\%) form the major components, while posters (10\%) and social-media app-style content (10\%) are included as more challenging yet practically important categories.

GPT-Image-2-Bench contains a total of 71 test samples and is constructed using a unified generation pipeline with gemini-3-flash-preview as the VLM/LLM backbone. For the document, DeepFake, and natural scene categories, we first leverage the VLM to describe samples from the existing OpenMMSec dataset, and then use GPT-Image-2 to regenerate corresponding images from these descriptions. For the \emph{AIGC} category, we sample prompts from DiffusionDB~\citep{wang2023diffusiondb}, which contains around 2 million real user prompts, and use GPT-Image-2 for image generation. For the \emph{poster} category, we employ the LLM to synthesize poster themes before generating images with GPT-Image-2. For the \emph{social-media app} category, the LLM is used to create platform-oriented content themes covering apps such as \textit{Xiaohongshu, Douyin, Twitter, WeChat, Weibo, Instagram, QQ,} and \textit{Telegram}, which are then rendered by GPT-Image-2. 

Overall, GPT-Image-2-Bench emphasizes both \emph{distributional diversity} and \emph{scenario realism}. The first four categories ensure broad coverage of mainstream generated-image content, while the poster and social-media categories further introduce text-rich, layout-complex, and style-driven samples that are closer to practical deployment settings. This benchmark therefore provides a targeted and challenging testbed for assessing model generalization on images generated by recent state-of-the-art generators. Examples from our GPT-Image-2-Bench are demonstrated in Figure~\ref{fig:vizgptbench}.

\section{Method}

\subsection{Model Architecture}
\begin{figure}[t]
    \centering
    \includegraphics[width=\linewidth]{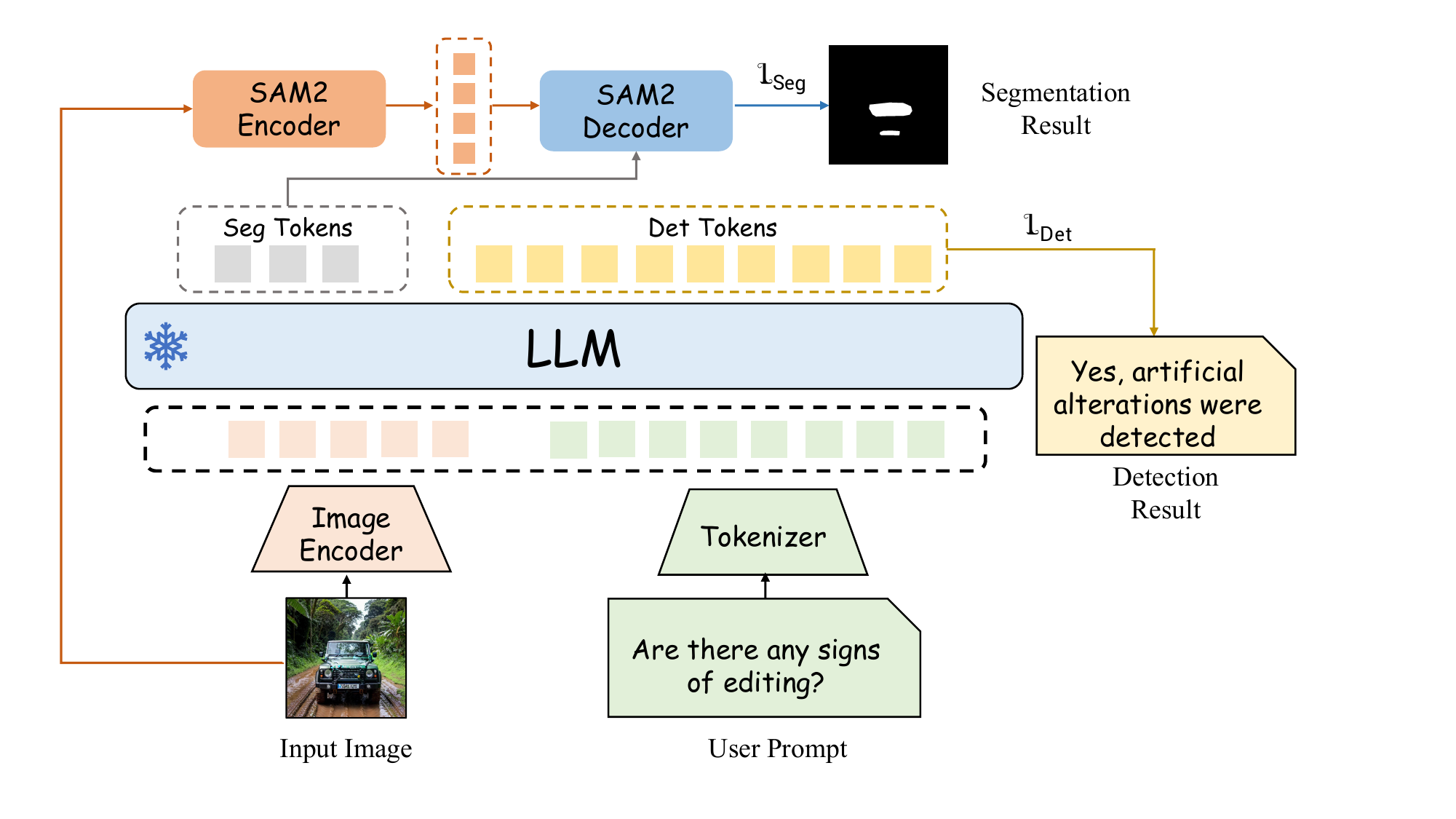} 
    \caption{\textbf{Overview of the DeFakerOne Architecture.} DeFakerOne consists of two main components: an MLLM-based perception-and-detection module and a SAM2-based segmentation module that leverages the aforementioned forensic information.}
    \label{fig:4_arch}
\end{figure}

Figure \ref{fig:4_arch} provides an overview of the DeFakerOne, which simultaneously integrates detection and segmentation for the anti-forgery task in a unified architecture. Specifically, DeFakerOne consists of two cascaded components: a Multimodal Large Language Model (MLLM)-based perception-and-detection module and a SAM2-based segmentation module. First, the MLLM is employed to perceive the input image and perform coarse-grained detection. Guided by carefully crafted dynamic VQA templates tailored for anti-forgery, the model analyzes the input image to provide a binary authenticity judgment and generates corresponding specific segmentation tokens for downstream fine-grained segmentation tasks. Furthermore, a segmentation module leverages this aforementioned forensic information to perform pixel-level analysis, producing segmentation masks that pinpoint the locations of detailed forgeries.

\subsection{MLLM-based Detection}
DeFakerOne selects InternVL2 ~\citep{chen2024internvl} as its MLLM backbone to address the critical need for advanced visual encoding in anti-forgery scenarios.
Unlike other semantics-oriented tasks, anti-forgery tasks prioritize localized low-level artifacts, such as visual generation artifacts and anomalous local details. Crucially, these forensic indicators often lack a deterministic correlation with the image's semantic content. Consequently, we propose reformulating the conventional simple binary real-or-fake detection problem into a dynamic Visual Question Answering (VQA) paradigm.
Specifically, as illustrated in Table \ref{tab:4_vqa_template}, rather than simply coupling the MLLM's output with a binary authenticity label, we have designed a series of dynamic VQA templates tailored for anti-forgery. By pairing the input question with either a positive or negative answer contingent on the binary ground-truth label, DeFakerOne effectively formulates the detection training task within a VQA framework. Adhering to the standard VQA training regimen serves a dual purpose: it guides the model in performing detection while simultaneously mitigating the degradation of its general answering capabilities. 


\begin{table}[]
\small
\centering
\caption{\textbf{The VQA Template in DeFakerOne.} To enhance annotation diversity and reduce reliance on image semantics, we propose reformulating the conventional binary real-or-fake detection problem into a dynamic Visual Question Answering (VQA) paradigm.}
\label{tab:4_vqa_template}
\begin{tabular}{ccc}

\hline
\rowcolor[HTML]{EFEFEF} 
\textbf{Questions}                                                                                     & \textbf{Positive Answers}                                                                               & \textbf{Negative Answers}                                                                                          \\ \hline
\begin{tabular}[c]{@{}c@{}}Are there any signs of\\  tampering in this image?\end{tabular}            & \begin{tabular}[c]{@{}c@{}}Yes, signs of human tampering\\  can be observed in the image.\end{tabular}      & \begin{tabular}[c]{@{}c@{}}No, no signs of human tampering\\  were found in the image.\end{tabular}                   \\ \hline
\begin{tabular}[c]{@{}c@{}}Does this image show \\ evidence of being manually altered?\end{tabular}   & \begin{tabular}[c]{@{}c@{}}Yeah, we detected traces of artificial\\  modification in the image.\end{tabular} & \begin{tabular}[c]{@{}c@{}}Not, this image shows no evidence\\  of any artificial modification.\end{tabular}           \\ \hline
\begin{tabular}[c]{@{}c@{}}Can you tell if this image\\  has been tampered with?\end{tabular}         & \begin{tabular}[c]{@{}c@{}}True, this image exhibits signs\\  of artificial manipulation.\end{tabular}       & \begin{tabular}[c]{@{}c@{}}Never, We did not find any signs\\  of tampering in the image.\end{tabular}                   \\ \hline
\begin{tabular}[c]{@{}c@{}}Are there any indications of\\  human editing in this image?\end{tabular}  & \begin{tabular}[c]{@{}c@{}}Sure, traces of human tampering\\  can be identified in the image.\end{tabular}   & \begin{tabular}[c]{@{}c@{}}None, no signs of artificial alteration\\  were detected in this image.\end{tabular}         \\ \hline
\begin{tabular}[c]{@{}c@{}}Does this image show signs\\  of having been processed?\end{tabular}       & \begin{tabular}[c]{@{}c@{}}Sure, the image displays evidence\\  of human modification.\end{tabular}          & \begin{tabular}[c]{@{}c@{}}Never, no signs of human tampering\\  were detected in the image.\end{tabular}                \\ \hline
\begin{tabular}[c]{@{}c@{}}Is there any evidence of human\\  manipulation in this image?\end{tabular} & \begin{tabular}[c]{@{}c@{}}Yeah, we observed traces of artificial\\  tampering in the image.\end{tabular}    & \begin{tabular}[c]{@{}c@{}}Not, this image does not appear to\\  have been tampered with by humans.\end{tabular}       \\ \hline
\begin{tabular}[c]{@{}c@{}}Can you identify whether\\  this image has been edited?\end{tabular}       & \begin{tabular}[c]{@{}c@{}}Yes, this image shows signs of\\  being manually modified.\end{tabular}          & \begin{tabular}[c]{@{}c@{}}Never,there is no evidence indicating\\  this image has been altered by humans.\end{tabular} \\ \hline
\begin{tabular}[c]{@{}c@{}}Has this image been\\  manually modified?\end{tabular}                     & \begin{tabular}[c]{@{}c@{}}True, tampering traces can\\  be identified in the image.\end{tabular}            & \begin{tabular}[c]{@{}c@{}}None, no traces of artificial\\  modification exist in the image.\end{tabular}               \\ \hline
\begin{tabular}[c]{@{}c@{}}Are there any signs\\  of editing in this image?\end{tabular}              & \begin{tabular}[c]{@{}c@{}}Yes, artificial alterations were \\ detected in the image.\end{tabular}          & \begin{tabular}[c]{@{}c@{}}No, there are no signs of\\  tampering in this image.\end{tabular}                         \\ \hline
\begin{tabular}[c]{@{}c@{}}Does this image appear\\  to have been tampered with?\end{tabular}         & \begin{tabular}[c]{@{}c@{}}Sure, the image shows visible\\  signs of digital tampering.\end{tabular}         & \begin{tabular}[c]{@{}c@{}}None, we did not find any evidence\\  of human modification in the image.\end{tabular}       \\ \hline
\end{tabular}
\end{table}

\subsection{Segmentation Module}

In the segmentation module, we employ an SA2VA-based~\citep{sa2va} architecture to achieve precise forgery region segmentation, effectively bridging the gap between semantic understanding and pixel-level localization. Specifically, for an input image, the LLM operates in a multi-task manner: while it outputs detection tokens for global classification, it simultaneously generates a set of specialized segmentation tokens. These tokens are not merely abstract representations; they encapsulate high-level semantic artifacts regarding the location and nature of potential manipulations, serving as dynamic prompts for the segmentation head.

Building on this, we leverage the SAM2 ~\citep{sam2} encoder to extract multi-scale hierarchical features from the input image, ensuring that both coarse semantic contexts and fine-grained texture details—critical for identifying subtle forgery artifacts—are preserved. The core of our approach lies in the feature fusion mechanism within the SAM2 decoder. Here, the LLM-derived segmentation tokens interact with the visual features via cross-attention mechanisms, effectively guiding the decoder to focus on suspicious regions identified during the detection phase. This semantic guidance allows the model to distinguish forged areas from complex backgrounds more robustly than traditional unsupervised methods. Finally, the decoder produces a high-resolution segmentation map indicating the forged regions, optimized using the segmentation losses to address the challenge of imbalanced foreground-background ratios typical in forgery detection tasks.

\subsection{Training Objective}
\label{sec:train}
Our framework integrates an MLLM with a segmentation layer, aiming to achieve both accurate classification and precise localization of image manipulations. Given an input image $I$ and a text query $T$, the MLLM first extracts semantic features to output the detection tokens $T_{Det}$ and segmentation tokens $T_{Seg}$, which explicitly define the image authenticity and provide the segmentation guidance, respectively. Subsequently, the image features and the segmentation tokens $T_{Seg}$ are fed into a dedicated segmentation layer $\mathcal{S}$ to generate a fine-grained segmentation mask $M$. The core of our approach lies in multi-stage joint training, which synergistically optimizes the Vision Encoder, the LLM, and the segmentation module to enhance the detection accuracy of manipulated regions.


To equip the model with the capability to determine authenticity and localize forged regions, we propose a joint optimization objective. Given a multimodal context sequence $\mathcal{X}$, the model is trained to predict discrete text tokens $\mathbf{x}$ for detection and continuous image masks $M$ for segmentation. The overall Supervised Fine-Tuning (SFT) objective function is formulated as:

\begin{equation}
    \mathcal{L}_{SFT} = \lambda_{txt} \mathcal{L}_{txt} + \lambda_{seg} \mathcal{L}_{seg},
\end{equation}
where $\mathcal{L}_{txt}$ is the standard autoregressive cross-entropy loss for the discrete text tokens,
defined as:
\begin{equation}
    \mathcal{L}_{txt} = - \sum_{i=1}^{N} \log P(x_i | \mathcal{X}_{<i}, I)
\end{equation}
Here, $N$ denotes the length of the target text sequence. 

Moreover, $\mathcal{L}_{seg}$ represents the segmentation loss used to optimize the resulting mask $M$, typically formulated as a combination of Binary Cross-Entropy (BCE) and Dice loss:
\begin{equation}
    \mathcal{L}_{seg} = \mathcal{L}_{BCE}(M, \hat{M}) + \mathcal{L}_{Dice}(M, \hat{M})
\end{equation}
where $\hat{M}$ is the ground-truth mask. The hyperparameters $\lambda_{txt}$ and $\lambda_{seg}$ are used to balance the text generation and visual segmentation objectives.

\subsection{Training Stages}

\paragraph{Stage 1: Paradigm Validation.}
In Stage 1, designated as domain convergence verification, we initialize from the InternVL2 checkpoint and conduct full-parameter supervised fine-tuning on 2 million curated forensic image-text pairs spanning the four target domains. This stage serves as a proof-of-concept to demonstrate that the model can simultaneously acquire discriminative features for diverse forgery patterns—ranging from GAN-generated artifacts to physics-inconsistent natural image manipulations—without catastrophic interference across task objectives. The relatively compact data scale and full-parameter update regime enable rapid adaptation of the pretrained vision-language representations to the forensic paradigm.

\paragraph{Stage 2: Capability Expansion and Scaling.} Stage 2 focuses on multi-domain capability enhancement through large-scale multi-task training. Building upon the Stage 1 checkpoint, we conduct comprehensive full-parameter supervised fine-tuning on an expanded corpus of 12.5 million samples, employing the AdamW optimizer for one epoch with a peak learning rate of $1\times10^{-5}$, warmup ratio of $0.05$, and batch size of $2$ samples per device, with linear decay and cosine annealing schedule. This stage employs balanced domain sampling to mitigate data imbalance across the four forensic categories, ensuring robust generalization without domain bias. The extended training corpus encompasses challenging edge cases, including adversarially perturbed forgeries and cross-domain composite manipulations, to enhance the model's forensic sensitivity.

\paragraph{Stage 3: Multi-Task Joint Refinement.} The final stage implements decoupled refinement with segmentation alignment, adopting modality-specific optimization strategies. For the language component, we apply LORA~\citep{hu2022lora} to the LLM layers with rank $r = 128$ and scaling factor $\alpha = 16$, maintaining the vision encoder and connector frozen, with a reduced learning rate of $1\times10^{-6}$ to prevent overfitting on high-level semantic descriptions while preserving acquired forensic reasoning patterns. Concurrently, we integrate a SAM-based segmentation module trained exclusively on document and natural image tampering datasets totaling 340K image-mask pairs, with the SAM backbone undergoing full-parameter fine-tuning using the AdamW optimizer at a peak learning rate of $1\times10^{-5}$, warmup ratio of $0.05$, and batch size of $2$, guided by domain-specific textual prompts for precise spatial delineation. This decoupled design ensures that the MLLM retains unified forgery detection capabilities without parameter explosion, while the SAM module specializes in fine-grained localization for structurally complex forgeries where pixel-accurate boundaries are critical for evidential validity.

\subsection{Inference Stage}
\label{sec:inference}

During inference, DeFakerOne performs image-level detection and pixel-level localization through a unified MLLM--SAM2 pipeline. Given an input image $I$ and a user query, the MLLM first generates detection-related textual responses and segmentation tokens. The former are used for authenticity prediction, while the latter are decoded by the SAM2 decoder to produce the localization mask.

\paragraph{Constrained-Vocabulary Tampering Detection.}
For image-level detection, we adopt a constrained-vocabulary scoring strategy to obtain a stable authenticity prediction. Given an image, we prompt the MLLM with a question such as ``Are there any signs of tampering in this image?'' Instead of directly relying on free-form generated answers, we analyze the first-token probability distribution over a curated vocabulary of eight response words:
\begin{equation}
    \mathcal{V}_{det} = \{\text{Yes}, \text{Yeah}, \text{True}, \text{Sure}, \text{No}, \text{Not}, \text{Never}, \text{None}\}.
\end{equation}
The logits of these tokens are normalized with a softmax operation:
\begin{equation}
    p(v|I,T) = \frac{\exp(z_v)}{\sum_{u \in \mathcal{V}_{det}} \exp(z_u)}, \quad v \in \mathcal{V}_{det},
\end{equation}
where $z_v$ denotes the first-token logit of token $v$. We then aggregate the probabilities of positive and negative response tokens to obtain the tampering score and authenticity score:
\begin{equation}
    S_{tamper} = \sum_{v \in \{\text{Yes}, \text{Yeah}, \text{True}, \text{Sure}\}} p(v|I,T),
\end{equation}
\begin{equation}
    S_{real} = \sum_{v \in \{\text{No}, \text{Not}, \text{Never}, \text{None}\}} p(v|I,T).
\end{equation}
Since $S_{tamper} + S_{real} = 1$, the final image-level prediction can be obtained by comparing the two scores, equivalently using a fixed decision boundary of $0.5$ for $S_{tamper}$. This avoids task-specific threshold tuning and provides a unified detection interface across different FIDL domains.

\paragraph{SAM2-based Forgery Localization.}
For pixel-level localization, DeFakerOne directly decodes the segmentation tokens generated by the MLLM. Specifically, the MLLM outputs segmentation tokens $T_{Seg}$ that serve as spatial prompts for the SAM2-based decoder. Given the image features extracted from the visual encoder and the segmentation guidance from $T_{Seg}$, the SAM2 decoder predicts the forgery mask:
\begin{equation}
    M = \mathcal{D}_{SAM2}(F_I, T_{Seg}),
\end{equation}
where $F_I$ denotes the visual feature representation of the input image and $\mathcal{D}_{SAM2}$ denotes the SAM2 decoder. The predicted mask $M$ highlights the manipulated regions at the pixel level. In this way, DeFakerOne uses the MLLM to provide both the image-level authenticity judgment and the high-level localization guidance, while SAM2 performs fine-grained mask decoding for accurate forgery localization.

\section{Results}

\begin{table}[t]
  \centering
  \caption{Image-level detection performance comparison on benchmarks. All results are averaged.}
  \label{tab:1_seg}
  \renewcommand{\arraystretch}{0.95}
  \resizebox{0.975\textwidth}{!}{
  \begin{tabular}{lcccccccccccc}
    \toprule
    \multirow{2}{*}{\textbf{Method}} 
    & \multicolumn{8}{c}{\textbf{Vision Models}} 
    & \multicolumn{4}{c}{\textbf{MLLMs}} \\
    \cmidrule(lr){2-9} \cmidrule(lr){10-13}
    & DTD & FFDN & Trufor & Mesorch & CDFA & Effort & DRCT & ForensicMOE 
    & FakeVLM & Veritas & Ivy-xDetector & \modelbase \\
    \midrule

    \multicolumn{13}{c}{\textbf{DeepFake (AUC)}} \\
    FF-c23~\citep{rossler2019faceforensics++} & 21.2 & 73.4 & 53.9 & 49.5 & \underline{94.2} & 92.5 & 66.7 & 47.7 & 83.4 & 4.9 & 71.7 & \textbf{99.4} \\
    FF-c40~\citep{rossler2019faceforensics++} & 20.8 & 79.1 & 49.2 & 49.8 & 79.7 & 80.4 & 59.2 & 64.9 & 68.8 & 4.4 & \underline{83.1} & \textbf{84.7} \\
    FF-DF~\citep{rossler2019faceforensics++} & 49.8 & 40.7 & 62.1 & 62.4 & \underline{99.1} & 99.0 & 78.8 & 46.0 & 63.9 & 3.1 & 73.4 & \textbf{99.8} \\
    FF-F2F~\citep{rossler2019faceforensics++} & 50.0 & 43.8 & 53.8 & 50.1 & \underline{93.8} & 92.0 & 59.3 & 36.4 & 60.1 & 3.9 & 67.3 & \textbf{99.2} \\
    FF-FS~\citep{rossler2019faceforensics++} & 50.2 & 42.6 & 47.0 & 39.3 & 96.5 & \underline{97.2} & 76.5 & 43.1 & 63.3 & 3.3 & 68.7 & \textbf{99.5} \\
    FF-NT~\citep{rossler2019faceforensics++} & 49.9 & 49.7 & 52.8 & 46.2 & \underline{87.4} & 81.8 & 52.5 & 34.9 & 61.7 & 2.8 & 66.6 & \textbf{98.8} \\
    CDFv1~\citep{li2020celeb} & 38.1 & 54.8 & 54.8 & 48.6 & 91.5 & \underline{90.7} & 43.1 & 19.9 & 72.2 & 7.8 & 53.0 & \textbf{99.9} \\
    CDFv2~\citep{li2020celeb} & 34.2 & 61.0 & 55.0 & 53.9 & \underline{91.4} & 88.2 & 48.7 & 16.2 & 81.9 & 7.9 & 53.6 & \textbf{99.9} \\
    DFD~\citep{dolhansky2020dfd} & 12.4 & 74.3 & 63.2 & 62.5 & \underline{93.2} & 92.2 & 77.4 & 81.9 & 88.9 & 7.4 & 77.9 & \textbf{98.7} \\
    DFDC~\citep{dolhansky2019deepfake} & 47.9 & 47.7 & 51.8 & 50.7 & \underline{84.2} & 82.2 & 61.2 & 30.6 & 60.1 & 4.5 & \underline{84.2} & \textbf{85.2} \\
    DFDCP~\citep{dolhansky2019deepfake} & 23.0 & 57.3 & 52.0 & 54.7 & \textbf{93.3} & 90.9 & 65.7 & 16.3 & 84.7 & 5.5 & 90.7 & \underline{91.0} \\
    Fsh~\citep{li2019faceshifter} & 50.0 & 43.1 & 54.0 & 52.5 & 77.7 & \underline{88.9} & 61.6 & 41.2 & 44.6 & 2.3 & 67.3 & \textbf{92.5} \\
    WDF~\citep{zi2020wilddeepfake} & 35.6 & 63.9 & 56.2 & 60.9 & 84.0 & \underline{85.9} & 63.4 & 21.1 & 63.9 & 6.3 & 82.1 & \textbf{93.0} \\
    ScaleDF~\citep{scaledf} & 22.4 & 73.7 & 44.2 & 53.4 & 76.0 & \underline{83.2} & 74.6 & 55.8 & 73.5 & 81.0 & 82.9 & \textbf{98.6} \\
    MFFI~\citep{miao2025mffi} & 46.1 & 55.2 & 51.0 & 55.8 & 76.5 & \underline{78.6} & 47.8 & 58.1 & 62.6 & 59.2 & 87.4 & \textbf{96.1} \\
    Avg & 36.8 & 57.4 & 53.4 & 52.7 & 87.9 & \underline{88.2} & 62.4 & 40.9 & 68.9 & 13.6 & 74.0 & \textbf{95.8} \\

    \midrule
    \multicolumn{13}{c}{\textbf{AIGC (Accuracy)}} \\
    ForenSynths~\citep{cnnspot} & 48.1 & 57.3 & 57.6 & 57.3 & 49.8 & \underline{74.8} & 68.6 & 71.5 & 71.3 & 69.5 & 77.0 & \textbf{77.2} \\
    DiffusionForensics~\citep{diffusion_detection} & 21.8 & 55.2 & 51.4 & 55.1 & 59.7 & 71.2 & 72.9 & 70.9 & 74.9 & 66.5 & \underline{93.5} & \textbf{99.5} \\
    GenImage~\citep{zhu2023genimage} & 43.8 & 55.1 & 53.4 & 55.0 & 59.0 & 90.5 & 87.8 & 94.3 & \underline{99.0} & 80.9 & 96.3 & \textbf{99.7} \\
    Chameleon~\citep{aide} & 36.7 & 37.9 & 41.2 & 37.8 & 55.2 & 63.1 & 51.6 & 58.9 & 62.9 & 59.6 & \underline{73.2} & \textbf{84.7} \\
    Fakeinversion~\citep{fakeinversion} & 83.8 & 38.4 & 36.7 & 38.3 & 42.3 & \underline{84.6} & 52.4 & 81.2 & 70.0 & 57.6 & 66.7 & \textbf{91.8} \\
    BFree-Online~\citep{bfree} & 36.6 & 67.8 & 60.7 & 67.8 & 38.3 & 55.9 & \underline{71.0} & 32.9 & \textbf{78.6} & 55.2 & 65.5 & 65.7 \\
    SynthWildx~\citep{SynthWildx} & 51.8 & 43.8 & 45.7 & 43.8 & 41.1 & 57.0 & 69.0 & 30.2 & 80.2 & \textbf{83.2} & \underline{81.9} & 80.1 \\
    EvalGEN~\citep{evalgen} & 24.7 & 40.2 & 17.6 & 40.2 & 33.2 & 35.8 & 77.7 & 18.9 & 92.1 & 90.4 & \underline{93.2} & \textbf{96.4} \\
    HydraFake~\citep{veritas} & 54.6 & 49.6 & 51.1 & 49.5 & 69.3 & 62.3 & 62.7 & 60.4 & 77.8 & \underline{87.2} & 71.7 & \textbf{92.8} \\
    Avg & 44.7 & 49.5 & 46.2 & 49.4 & 49.8 & 66.1 & 68.2 & 57.7 & 78.5 & 72.2 & \underline{79.9} & \textbf{87.5} \\

    \midrule
    \multicolumn{13}{c}{\textbf{Doc (Accuracy)}} \\
    DocTamper\_FCD~\citep{qu2023towards} & \underline{73.2} & 73.1 & 68.4 & 64.2 & 52.3 & 60.1 & 77.0 & 38.9 & 1.9 & 39.4 & 25.3 & \textbf{92.8} \\
    DocTamper\_SCD~\citep{qu2023towards} & 69.8 & 58.7 & 67.7 & 53.9 & 39.8 & 26.7 & \underline{71.8} & 24.6 & 1.0 & 31.2 & 25.1 & \textbf{99.8} \\
    DocTamper\_TestingSet~\citep{qu2023towards} & 64.3 & 57.8 & 60.9 & 54.2 & 46.9 & 37.1 & \underline{66.8} & 29.8 & 29.5 & 39.0 & 33.9 & \textbf{99.6} \\
    TextForensicsReasoning~\citep{qu2026textshield} & 43.0 & 51.3 & 51.8 & 52.7 & 51.0 & 50.5 & 52.2 & 46.9 & 52.6 & 52.0 & \underline{54.6} & \textbf{99.6} \\
    Tampered IC13~\citep{wang2022tpic} & \textbf{75.3} & \underline{61.9} & 58.8 & 51.5 & 48.0 & 19.0 & 48.9 & 30.5 & 25.3 & 38.2 & 22.8 & 54.9 \\
    OSTF~\citep{qu2025revisiting} & \underline{60.3} & 47.0 & 59.4 & 51.2 & 57.7 & 55.9 & 57.1 & 57.9 & 57.0 & 56.8 & 58.0 & \textbf{74.9} \\
    RTM~\citep{luo2025rtm} & 45.1 & 51.6 & 59.8 & 47.2 & 47.8 & 38.2 & \underline{63.0} & 33.6 & 32.5 & 42.3 & 33.7 & \textbf{90.0} \\
    Avg & 61.6 & 57.3 & 61.0 & 53.6 & 49.1 & 41.1 & \underline{62.4} & 37.5 & 28.5 & 42.7 & 36.2 & \textbf{87.4} \\

    \midrule
    \multicolumn{13}{c}{\textbf{Nature (AUC)}} \\
    CASIAv1~\citep{CASIA_2013} & 46.8 & 52.5 & 93.9 & \underline{95.0} & 63.9 & 61.9 & 52.7 & 7.7 & 4.2 & 3.2 & 45.8 & \textbf{96.3} \\
    COVERAGE~\citep{Coverage_2016} & 51.5 & 40.5 & 72.3 & \textbf{76.1} & 53.6 & 49.2 & 52.9 & 50.6 & 50.0 & 0.4 & 55.5 & \underline{75.9} \\
    Columbia~\citep{Columbia_2006} & 62.3 & 2.20 & \underline{99.4} & \textbf{99.8} & 88.0 & 49.2 & 48.2 & 81.7 & 52.6 & 2.5 & 87.6 & 87.7 \\
    NIST16~\citep{NIST16_2019} & 56.0 & 38.7 & 59.4 & 66.5 & 61.6 & 73.4 & \underline{66.7} & 3.9 & 58.7 & 1.2 & 55.4 & \textbf{85.1} \\
    CocoGlide~\citep{trufor2023} & 51.9 & 29.6 & 67.1 & \underline{71.9} & 67.8 & 68.2 & 72.3 & 19.2 & 66.0 & 3.3 & 65.3 & \textbf{76.0} \\
    Autosplice~\citep{jia2023autosplice} & 42.9 & 34.3 & 64.5 & 67.6 & \underline{77.7} & 82.0 & 75.5 & 37.4 & 67.2 & 4.0 & 40.8 & \textbf{99.9} \\
    OpenSDI~\citep{opensdi} & 35.2 & 42.5 & 57.9 & 57.1 & 52.4 & 77.7 & \underline{84.9} & 5.0 & 73.5 & 1.9 & 69.5 & \textbf{98.6} \\
    DSO-1~\citep{dso-1} & \underline{71.0} & 35.0 & 64.5 & 69.0 & 64.1 & 55.9 & 53.6 & 23.2 & 60.5 & 0.5 & 15.1 & \textbf{82.4} \\
    DEFACTO-12k~\citep{defacto_2019} & 55.1 & 37.8 & \underline{64.3} & 63.9 & 53.6 & 53.5 & 55.3 & 1.7 & 50.0 & 0.9 & 30.9 & \textbf{78.3} \\
    Avg & 52.5 & 34.8 & 71.5 & \underline{74.1} & 64.7 & 63.4 & 62.5 & 25.6 & 53.6 & 2.0 & 51.8 & \textbf{86.7}  \\

    \bottomrule
  \end{tabular}
  }
\end{table}
\begin{table}
  \caption{Performance (Accuracy) comparison with baseline methods on OpenMMsec; \textbf{Bold} indicates the best performance; \underline{Underline} indicates the second-best performance.}
  \label{tab:2_mmsec}
  \centering
  \resizebox{0.85\linewidth}{!}{%
  \begin{tabular}{lcccccc}
    \toprule
    \textbf{Method} & \textbf{Venue} & \textbf{DeepFake} & \textbf{AIGC} & \textbf{IMDL} & \textbf{Doc} & \textbf{Avg} \\
    \midrule
    Resnet~\citep{Resnet_2016} & CVPR'16 & 67.8 & 74.2 & 77.3 & 71.7 & 72.7 \\
    EfficientNet~\citep{efficientnet} & ICML'19 & 43.3 & 63.0 & 51.4 & 61.9 & 54.9 \\
    CapsuleNet~\citep{sabour2017dynamic} & NeurIPS'17 & 62.0 & 75.4 & 77.2 &  72.3 & 71.7 \\
    SegFormer~\citep{SegFormer_2021} & NeurIPS'21 & 80.7 & 85.9 & 81.7 & \underline{73.4} & \underline{80.4} \\
    Swin~\citep{Swin_2021} & ICCV'21 & 79.0 & 85.4 & 82.9 & 72.0 & 79.8 \\
    Trufor~\citep{trufor2023} & CVPR'23 & 72.2 & 82.5 & 80.5 & 72.3 & 76.9 \\
    UnivFD~\citep{univfd} & CVPR'23 & 54.2 & 79.8 & 70.9 & 62.7 & 66.9 \\
    Effort~\citep{effort} & ICML'25 & \underline{85.0} & 81.9 & \underline{83.7} & 69.6 & 80.1 \\
    Mesorch~\citep{zhu2025mesorch} & AAAI'25 & 75.7 & 81.4 & 79.9 & 72.7 & 77.4 \\
    CO-SPY~\citep{co-spy} & CVPR'25 & 72.2 & 83.3 & 76.0 & 70.0 & 75.4 \\
    \midrule
    FakeVLM~\citep{fakeclue} & NeurIPS'25 & 75.8 & \underline{86.6} & 66.2 & 14.0 & 60.7 \\
    Veritas~\citep{veritas} & ICLR'26 & 83.4 & 72.8 & 66.8 & 32.1 & 63.8 \\
    FakeShield~\citep{xu2024fakeshield} & ICLR'25 & 72.2 & 73.2 & 59.5 & 68.5 & 68.4 \\
    Ivy-xDetector~\citep{zhang2025ivyfakeunifiedexplainableframework} & Arxiv'25 & 56.3 & 83.1 & 64.2 & 17.7 & 55.3 \\
    \modelbase (Ours) & - & \textbf{89.5} & \textbf{96.4} & \textbf{91.1} & \textbf{90.1} & \textbf{91.8} \\
    \bottomrule
    \end{tabular}
    }
\end{table}

\begin{table*}[t]
\centering
\small
\setlength{\tabcolsep}{6.5pt}
\caption{Pixel-level forgery localization performance on segmentation benchmarks using binary F1 score (\%). The upper block reports Document benchmarks with document-oriented baselines, and the lower block reports Nature benchmarks with nature-oriented baselines. Unreported results are denoted by ``--''.}
\label{tab:2_seg}
\begin{tabular}{lcccc}
\toprule
\multicolumn{5}{c}{\textbf{Document Benchmarks}} \\
\midrule
Method & DTD & CAFTB & TIFDM & \modelbase \\
\midrule
DocTamperFCD~\citep{qu2023towards}  & {\ul 68.6}    & 29.2       & 9.0  & \textbf{80.8} \\
DocTamperSCD~\citep{qu2023towards}  & {\ul 73.9}    & 37.7       & 25.7 & \textbf{74.9} \\
DocTamperTest~\citep{qu2023towards} & \textbf{80.3} & 32.8       & 25.9 & {\ul 73.3}    \\
T-SROIE~\citep{wang2022tsroie}       & \textbf{92.1} & {\ul 91.7} & 89.4 & 81.7          \\
Tampered IC13~\citep{wang2022tpic} & 83.4          & {\ul 83.9} & 79.7 & \textbf{90.5} \\
OSTF~\citep{qu2025revisiting}          & 56.3          & {\ul 64.8} & 54.1 & \textbf{90.6} \\
RTM~\citep{luo2025rtm}           & 17.2          & {\ul 24.9} & 5.9  & \textbf{59.1} \\
\midrule
Avg           & {\ul 67.4}    & 52.1       & 41.4 & \textbf{78.7} \\
\midrule
\multicolumn{5}{c}{\textbf{Nature Benchmarks}} \\
\midrule
Method & MVSS-Net & TruFor & Mesorch & \modelbase \\
\midrule
CASIAv1~\cite{CASIA_2013}   & 43.5 & 69.2       & \textbf{84.0} & {\ul 78.5}    \\
COVERAGE~\cite{Coverage_2016}  & 45.4 & 52.2       & {\ul 58.6}    & \textbf{71.7} \\
Columbia~\cite{Columbia_2006}  & 78.1 & {\ul 85.9} & \textbf{89.0} & 80.9          \\
NIST16~\cite{NIST16_2019}    & 29.4 & 34.8       & {\ul 39.2}    & \textbf{65.6} \\
CocoGlide~\citep{trufor2023} & {\ul29.1}   & 20.5         & 16.2      & \textbf{62.0} \\
AutoSplice~\citep{jia2023autosplice}& 29.4   & {\ul39.3}         & 24.9      & \textbf{45.7} \\
\midrule
Avg & 42.5 & 50.3 & {\ul52.0} & \textbf{67.4} \\
\bottomrule
\end{tabular}
\end{table*}

\begin{table}[!t]
\caption{
Average robustness analysis on OpenMMsec~\citep{du2026SICA}. 
This table only reports the averaged results over perturbation strengths, while the complete robustness table is provided in Table~\ref{tab:3_robust_full}.
}
  \label{tab:3_robust}
  \centering
  \resizebox{0.95\linewidth}{!}{%
  \begin{tabular}{l|ccccccc}
    \toprule
    \textbf{Perturbation} 
    & \textbf{FFDN} 
    & \textbf{Mesorch} 
    & \textbf{Effort} 
    & \textbf{ForensicsAdapter} 
    & \textbf{ForensicMOE} 
    & \textbf{FakeShield} 
    & \textbf{\modelbase} \\
    \midrule
    Gaussian Blur    & 47.63 & 49.45 & 56.71 & 55.76 & 54.11 & 64.63 & \textbf{79.46} \\
    Brightness       & 44.78 & 51.06 & 57.53 & 56.10 & 54.82 & 63.48 & \textbf{70.60} \\
    Contrast         & 43.65 & 51.12 & 58.27 & 56.22 & 55.53 & 63.86 & \textbf{76.26} \\
    JPEG Compression & 39.93 & 52.00 & 58.73 & 56.78 & 51.67 & 62.77 & \textbf{76.16} \\
    Noise            & 48.03 & 48.18 & 54.62 & 52.77 & 53.62 & 63.54 & \textbf{65.32} \\
    Resize           & 44.19 & 50.54 & 58.69 & 57.02 & 53.47 & 51.30 & \textbf{69.23} \\
    Saturation       & 40.69 & 51.69 & 58.43 & 57.30 & 55.77 & 64.30 & \textbf{81.85} \\
    \bottomrule
  \end{tabular}
  }
\end{table}


\subsection{Performance Comparison Across FIDL Domains}
\begin{table}[t]
\begin{minipage}{0.56\textwidth}
\centering
\captionof{table}{Ablation Study on Forensic Domains and Tasks.}
\label{tab:single_vs_multi_domain}
\resizebox{\linewidth}{!}{
\begin{tabular}{lccccc}
\toprule
\textbf{Model} & \textbf{Doc} & \textbf{IMDL} & \textbf{Deepfake} & \textbf{AIGC} & \textbf{Avg.} \\
\midrule
DeFakerOne (stage3. Multi-Task) & \textbf{92.5} & \underline{89.7} & \underline{89.0} & \textbf{96.8} & \textbf{92.0} \\
\midrule
DeFakerOne (stage2. Data-Scaling) & \underline{90.1} & \textbf{91.1} & \textbf{89.5} & \underline{96.4} & \underline{91.8} \\
\midrule
DeFakerOne (stage1. Multi-Domain) & 89.7 & 54.5 & 77.4 & 85.9 & 76.8 \\
\midrule
Nature Specialized   & 66.6 & 74.6 & 64.3 & 74.5 & 70.0 \\
AIGC Specialized     & 13.9 & 53.2 & 28.3 & 92.9 & 47.1 \\
Deepfake Specialized & 13.8 & 47.6 & 88.7 & 45.8 & 49.0  \\
Doc Specialized      & 89.5 & 55.5 & 29.2 & 44.3 & 54.6 \\
\bottomrule
\end{tabular}
}
\end{minipage}
\hfill
\begin{minipage}{0.40\textwidth}
\centering
\captionof{table}{Decoding hyperparameters.}
\label{tab:5_tts}
\resizebox{\linewidth}{!}{
\begin{tabular}{c|cc @{\hspace{1.2em}} c|cc}
\toprule
\multicolumn{3}{c}{\textbf{(a) Seed}} & \multicolumn{3}{c}{\textbf{(b) Temperature}} \\
\cmidrule(r){1-3} \cmidrule(l){4-6}
\textbf{Seed} & \textbf{ACC} & \textbf{F1} & \textbf{Temp} & \textbf{ACC} & \textbf{F1} \\
\midrule
42   & 91.8 & 93.4 & 0.1 & 91.8 & 93.4 \\
1024 & 91.7 & 93.5 & 0.5 & 91.8 & 92.5 \\
8192 & 91.8 & 93.5 & 0.9 & 91.7 & 93.5 \\
\bottomrule
\end{tabular}
}
\end{minipage}
\end{table}

\paragraph{Comparison with FIDL-domain MLLMs.}
As shown in Table~\ref{tab:1_seg}, DeFakerOne~consistently achieves SOTA performance across multiple domains. Veritas demonstrates slightly weaker results in terms of AUC. The overall performance of FakeVLM and Ivy-xDetector is limited, which can be attributed to the lack of training data in the doc domain. Notably, on the Chameleon task, DeFakerOne~achieves the best result of 84.7\%, substantially surpassing Ivy-xDetector (73.2\%).

\paragraph{Comparison with Vision Models.}
As shown in Table~\ref{tab:1_seg}, small vision models show limited cross-benchmark generalization, while MLLM-based detectors generally achieve stronger performance. However, FakeVLM and Ivy-xDetector remain weak in the Document domain, highlighting the importance of document-oriented training data. In contrast, DeFakerOne achieves the best average performance across all evaluated domains and obtains top results on most benchmarks.

\paragraph{Cross-domain Analysis.}
As shown in Table~\ref{tab:2_mmsec}, we compare model performance across four domains on the OpenMMsec dataset. Small vision models outperform MLLM-based detectors that have not been trained on OpenMMsec, indicating limited cross-domain generalization for such models. FakeVLM and Ivy-xDetector achieve below 20\% accuracy in the doc domain, which can be attributed to the lack of Document category samples in their training data, underscoring the critical role of data distribution. In contrast, DeFakerOne~achieves the highest performance, reaching 91.8\%.

\paragraph{Segmentation Performance Comparison.}
As shown in Table~\ref{tab:2_seg}, DeFakerOne demonstrates strong segmentation performance across both document and natural image forgery benchmarks. 
For document forgery, it achieves the best pixel-level F1 on five out of seven benchmarks and obtains the highest average score of 78.7\%, clearly surpassing the second-best method DTD (67.4\%). 
This indicates its strong ability to localize subtle text manipulations in complex document images. 
For natural images, DeFakerOne achieves the best average F1 of 67.4\%, outperforming the second-best method Mesorch (52.0\%). 
It ranks first on four out of six benchmarks, including COVERAGE, NIST16, CocoGlide, and AutoSplice, while remaining competitive on CASIAv1 and Columbia. 
These results show that DeFakerOne provides reliable forgery localization across both document-oriented and natural-image manipulation scenarios.

\paragraph{Robustness Analysis.} As shown in Table~\ref{tab:3_robust}, we evaluate model robustness under a diverse set of common image perturbations, including Gaussian blur, brightness/contrast adjustment, JPEG compression, additive noise, resizing, and saturation variation. Overall, existing baselines exhibit noticeable performance degradation as perturbation strength increases, indicating limited robustness to distribution shifts.

Under Gaussian blur, most methods suffer from a monotonic decline as blur intensity increases, reflecting sensitivity to high-frequency information loss. DeFakerOne~maintains a clear margin across all levels and achieves the highest average accuracy (79.46\%), demonstrating strong resilience to spatial detail degradation. A similar trend is observed for brightness and contrast variations, where baseline methods show instability under extreme conditions, while DeFakerOne remains consistently superior, indicating better invariance to illumination changes.

For compression and noise perturbations, which simulate real-world transmission and acquisition artifacts, baseline methods again show significant performance drops. In contrast, DeFakerOne~retains robust performance, suggesting that it captures more stable and semantically meaningful features rather than relying on brittle low-level cues. Notably, under resizing operations, where spatial resolution changes can disrupt feature alignment, most baselines degrade substantially, whereas DeFakerOne~continues to outperform all competitors, highlighting its robustness to scale variations.

\paragraph{Decoding Hyperparameters.}To assess the inference stability of DeFakerOne, we conduct comprehensive robustness evaluations on decoding hyperparameters, with results presented in Table \ref{tab:5_tts}. Regarding random seed variations, we evaluate three distinct seeds (42, 1024, and 8192) and observe virtually identical performance across all configurations, with accuracy stable at 91.7–91.8\% and F1 scores consistently ranging from 93.4\% to 93.5\%. This negligible variance demonstrates that our unified training paradigm effectively eliminates sensitivity to initialization stochasticity, ensuring reproducible predictions in deployment scenarios.
We further examine temperature scaling, a critical parameter controlling prediction diversity in generative models. Across temperatures ranging from 0.1 (highly deterministic) to 0.9 (increased stochasticity), DeFakerOne maintains remarkable stability with accuracy fluctuating within merely 0.1 percentage points (91.7–91.8\%) and F1 scores remaining stable at 92.5–93.5\%. The minimal performance degradation at elevated temperatures, where one might expect increased prediction variance, substantiates that the model has learned robust forensic decision boundaries rather than relying on brittle spurious correlations. These results collectively establish that DeFakerOne delivers consistent, reliable outputs across diverse operational configurations, a critical prerequisite for forensic applications where verdict reproducibility carries legal and evidentiary significance.

Across all perturbation categories, the average performance of baseline methods typically falls within a limited range, while DeFakerOne~consistently operates at a significantly higher level. This consistent advantage across heterogeneous corruptions indicates that DeFakerOne~learns more generalized and perturbation-invariant representations. Overall, these results demonstrate that DeFakerOne~not only excels in clean settings but also provides strong robustness under realistic and challenging conditions, making it more reliable for practical deployment.

\begin{figure}[t]
    \centering
    \includegraphics[width=0.72\linewidth]{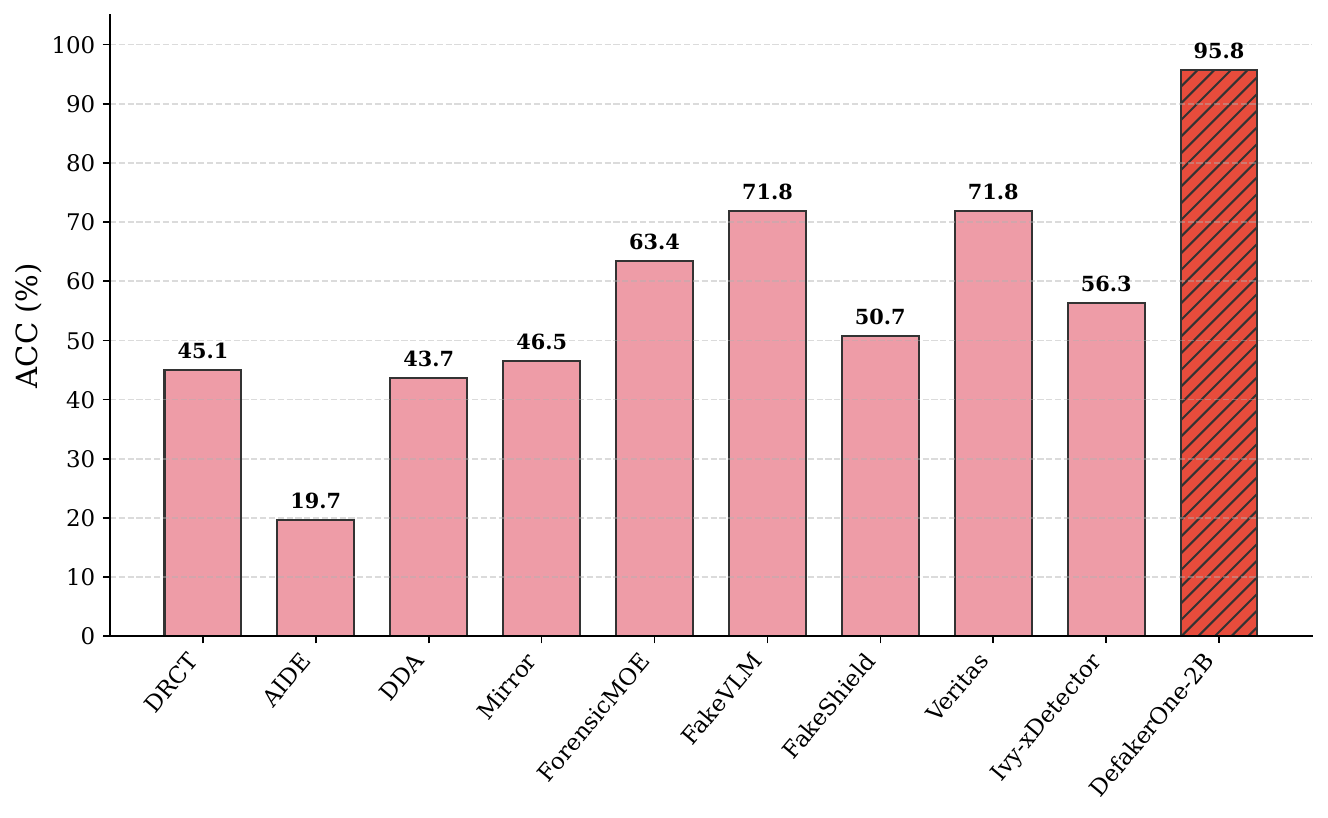}
    \caption{
    Accuracy comparison on GPT-Image-2-Bench. GPT-Image-2 generates high-quality images with fewer obvious synthesis artifacts, posing a challenging distribution shift for existing detectors. DeFakerOne achieves the best performance among all compared methods.
    }
    \label{fig:gpt-image-2-test}
\end{figure}

\paragraph{Performance on GPT-Image-2-Bench.}
As shown in Figure~\ref{fig:gpt-image-2-test}, GPT-Image-2 introduces a challenging distribution shift for existing fake image detectors. Compared with earlier diffusion-based or domain-specific generators, GPT-Image-2 produces images with stronger semantic coherence, cleaner local textures, and fewer low-level synthesis artifacts, making conventional artifact-driven detectors less reliable. As a result, several prior methods achieve only moderate accuracy on this benchmark, e.g., DRCT, DDA, Mirror, and FakeShield remain below 51\%, while stronger MLLM-based or forensic-specialized baselines such as FakeVLM, Veritas, and ForensicMOE perform better but still leave a large gap. In contrast, DeFakerOne achieves 95.77\% accuracy, indicating that our model generalizes more effectively to high-quality images generated by the latest image synthesis model. This result suggests that detecting GPT-Image-2 images requires not only low-level artifacts recognition, but also higher-level forensic reasoning over semantic consistency, layout plausibility, and cross-region visual evidence.

\subsection{Result Analysis}
\begin{figure}[t]
    \centering
    \includegraphics[width=0.92\linewidth]{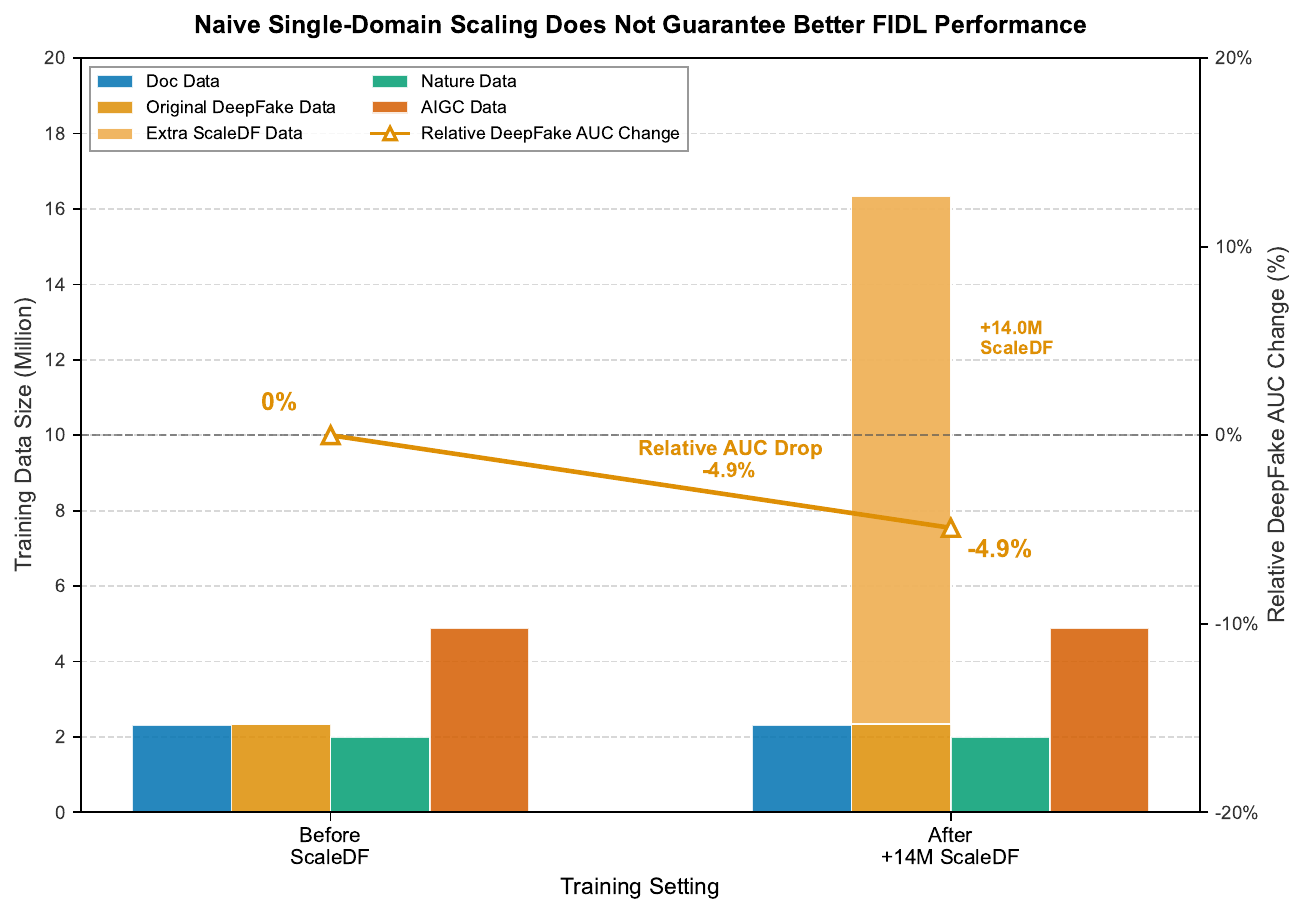}
    \caption{Extreme single-domain scaling in the DeepFake domain. Although adding 14M ScaleDF samples substantially increases the DeepFake training data size, the relative DeepFake AUC decreases by 4.9\%, suggesting that naive single-domain scaling does not guarantee better FIDL performance.}
    \label{fig:deepfake_scaling}
\end{figure}

\subsubsection{Data Scaling Does Not Guarantee FIDL Performance.}We first conduct an extreme single-domain scaling experiment to examine whether increasing the data scale of one domain can improve the performance of that target domain. Specifically, in the DeepFake domain, we expand the original 2.336M face forgery training samples by adding about 14M ScaleDF samples, increasing the DeepFake-related training data to about 16.336M samples. If the model could straightforwardly benefit from single-domain data scaling, this expansion should lead to a clear improvement in DeepFake performance. However, the results do not support this assumption. As shown in Figure~\ref{fig:deepfake_scaling}, after adding the large-scale ScaleDF data, the performance on open-source DeepFake benchmarks does not continue to improve, but instead drops by about 4.9\% compared with the original model. This indicates that even for the target domain itself, continuously increasing the scale of single-domain data does not necessarily lead to better performance. The model may quickly reach a performance plateau, or even degrade due to distribution shift, sample redundancy, or imbalanced data composition. In other words, single-domain FIDL performance does not follow a simple ``more data is better'' scaling law. Therefore, FIDL data scaling should not be understood as unconstrained scaling up of a single domain. For DeFakerOne, a data scale of tens of millions is not sufficient by itself. The key lies in the ratio, quality, and distributional complementarity among different domains. This observation supports our data-centric view: building a FIDL foundation model cannot rely on simply stacking the largest possible amount of data, but requires systematic analysis of data composition, transfer, and interference across domains.

\subsubsection{Operation-Level Artifacts Drive Transfer and Interference}

\begin{figure}[t]
    \centering
    \includegraphics[width=0.98\linewidth]{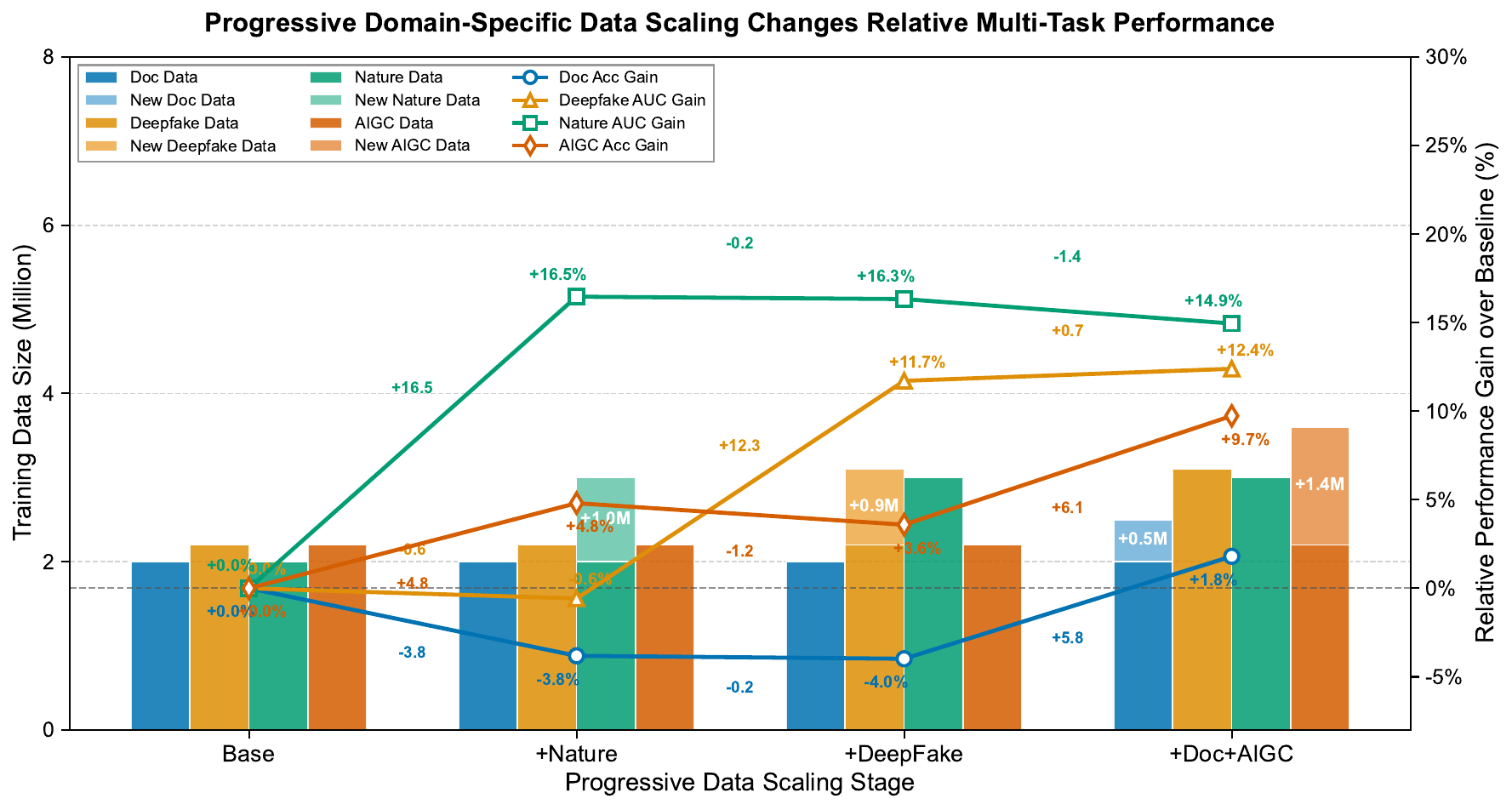}
    \caption{
    Progressive domain-specific data scaling and its effect on relative multi-task FIDL performance.
    The bars show the training data size of each domain, where the light-colored segments denote newly added data at each scaling stage.
    The curves report the relative performance gain over the 8M baseline.
    Target-domain supplementation generally improves the corresponding domain, but may also introduce transfer or interference effects on other domains.
    }
    \label{fig:data_scaling}
\end{figure}

We further analyze how progressive domain-specific data supplementation affects unified FIDL performance. As shown in Figure~\ref{fig:data_scaling}, the effect of adding domain-specific data is not simply positive or negative. Instead, it exhibits a mixed transfer--interference behavior across domains.

Specifically, after adding Nature-related data, the Nature performance increases by about 16.5\%, and the AIGC performance also improves. However, Doc and DeepFake decline at the same stage. This suggests that Nature data not only benefits its target domain, but can also transfer to certain non-target domains with compatible artifacts patterns.

\begin{table}[t]
    \centering
    \caption{
    Operation-level changes after adding AIGC+Doc data over the Nature-only setting.
    ``Nature Avg.'' denotes the average performance over all Nature benchmarks, while CocoGlide, AutoSplice, and OpenSDI report changes on individual subsets.
    }
    \label{tab:operation_transfer}
    \begin{tabular}{lcccc}
        \toprule
        Setting & Nature Avg. & CocoGlide & AutoSplice & OpenSDI \\
        \midrule
        Gain over Nature-only & -1.6\% & +9.48\% & +20.42\% & +13.83\% \\
        \bottomrule
    \end{tabular}
\end{table}

These results reveal an apparent contradiction: adding one domain can sometimes benefit another domain, but can also suppress others. Therefore, the transfer behavior cannot be fully explained by coarse domain labels such as Nature, AIGC, DeepFake, and Document. To understand this phenomenon, we further analyze the results at a finer granularity. As shown in Table~\ref{tab:operation_transfer}, although adding AIGC-related data may reduce the overall Nature average, some Nature subsets involving AIGC-style local manipulation, generative editing, or semantic completion still improve.

This indicates that cross-domain transfer and interference are mainly determined by operation-level artifacts similarity. Datasets with similar manipulation mechanisms, such as generative texture bias, semantic completion traces, blending inconsistency, boundary artifacts, face-swapping traces, or text replacement artifacts, can benefit from each other even if they belong to different macro domains. Conversely, incompatible artifacts patterns may introduce interference. Therefore, FIDL data should be organized not only by domains but also according to manipulation methods.

\subsubsection{Balanced Data Recomposition Is the Key to Unified FIDL}

After identifying the transfer--interference behavior caused by operation-level artifacts similarity, we further ask how to stabilize unified FIDL training under such cross-domain interactions. The above analysis shows that targeted data supplementation can introduce both transfer and interference, depending on the operation-level artifacts similarity. This further suggests that unified FIDL cannot be achieved by simply maximizing the data scale of a single domain. Even when the target domain is improved, the model may become biased toward the newly emphasized artifacts patterns, thereby weakening its sensitivity to other forgery traces. Therefore, beyond operation-aware data organization, FIDL also requires balanced data recomposition across domains.

Following this observation, we further re-compose the training data to restore multi-domain balance. As shown in the final stage of Figure~\ref{fig:data_scaling}, after improving Nature and DeepFake performance, we continue to supplement the Document and AIGC datasets. This shifts the training process from target-domain enhancement to a more balanced four-domain composition. The results show that the previously weakened domains are effectively recovered: the average Doc performance improves by about 6.0\%, and the average AIGC performance improves by about 5.9\%. Meanwhile, DeepFake remains stable with a slight improvement, and Nature only shows a small fluctuation of about 1.2\%. This indicates that supplementing the interfered domains can restore their performance while largely preserving the gains obtained in earlier stages.

Overall, the average performance across the four major domains improves by about 9.6\%. Importantly, this gain is not caused by a simple monotonic increase in total data size. Instead, it follows a data evolution process of ``target-domain enhancement -- cross-domain interference -- supplementation of weakened domains -- global re-balancing.'' Finally, DeFakerOne forms a data composition where Document, DeepFake, Nature, and AIGC are maintained at comparable scales, achieving the best overall performance under the current experimental setting. These results suggest that the key to unified FIDL is not allowing one domain to dominate, but re-composing data under a balanced and operation-aware multi-domain scale, so that detection, localization, and cross-domain generalization can be jointly maintained.

\subsubsection{Different FIDL Domains Require Different Learning and Supervision Granularity}

Beyond data scale and domain composition, we further explore whether different FIDL domains require different supervision granularity. The above analysis shows that hard domains such as Nature and Doc are sensitive to data composition. We argue that this is also related to the granularity of their forensic cues. DeepFake and AIGC are relatively easier to learn from image-level labels: DeepFake manipulations are usually concentrated on face regions, while AIGC images often contain more stable global generation traces. In contrast, Nature and Doc forgeries are more fine-grained. Nature forgeries often involve local splicing, copy-move, object removal, or generative local editing, while document forgeries may only modify small text regions, numbers, seals, or local layouts. In these cases, key forensic cues are local and weak, and may be overwhelmed by global semantic representations.

\begin{table}[t]
    \centering
    \caption{
    Effect of segmentation supervision on Nature benchmarks.
    The table reports the AUC gain brought by adding segmentation supervision over classification-only training.
    }
    \label{tab:seg_supervision}
    \resizebox{0.82\linewidth}{!}{
    \begin{tabular}{lcccccccc}
        \toprule
        Supervision & Coverage & Columbia & NIST16 & CocoGlide & Autosplice & DSO-1 & Avg. \\
        \midrule
        Gain (\%) & +1.1 & +2.0 & +10.7 & +2.9 & +0.4 & +2.5 & +3.3 \\
        \bottomrule
    \end{tabular}
    }
\end{table}

Based on this observation, we introduce segmentation supervision to examine whether finer-grained supervision can improve these hard local-manipulation scenarios. As shown in Table~\ref{tab:seg_supervision}, joint classification and segmentation training improves not only localization ability, but also image-level classification performance. Compared with classification-only supervision, adding segmentation masks improves the average AUC by 3.3 percentage points on representative Nature local-manipulation benchmarks, with especially large gains on NIST16. This suggests that pixel-level masks help the model focus on manipulation boundaries, local residuals, and regional inconsistencies, which are difficult to learn from image-level labels alone.

Therefore, unified FIDL should consider not only data scale and domain composition, but also supervision granularity. Current forgery attacks are increasingly fine-grained, targeting text, boundaries, local textures, or small objects. If the defense model only relies on coarse image-level real/fake labels, it may fail to capture these weak local traces. This ``fine-grained attack vs. coarse-grained defense'' asymmetry explains why challenging domains such as Nature and Doc benefit from pixel-level or region-level supervision. Fine-grained supervision is therefore useful not only for localization, but also for strengthening image-level detection in hard FIDL domains.

\begin{table}[t]
    \centering
    \small
    \setlength{\tabcolsep}{4.5pt}
    \caption{
    Relative performance of different VLM backbones for unified FIDL.
    InternVL2-2B is used as the baseline.
    }
    \label{tab:backbone_relative}
    \begin{tabular}{lccccc}
        \toprule
        Backbone & DeepFake & Document & AIGC & Nature & Avg. \\
        \midrule
        InternVL2-2B    & --      & --       & --      & --      & -- \\
        InternVL3.5-2B  & -0.6\%  & -13.2\% & -1.2\% & -2.0\% & -4.3\% \\
        Qwen3-VL-2B     & -0.4\%  & -10.4\% & +5.0\% & +1.5\% & -1.1\% \\
        \bottomrule
    \end{tabular}
\end{table}

\subsubsection{Original Resolution Artifacts Preservation Is Crucial for Unified FIDL}

Beyond data scale and domain composition, we further analyze the impact of different multimodal backbones on unified FIDL. Specifically, we use InternVL2-2B as the baseline and compare it with newer backbones, including InternVL3.5-2B, Qwen3-VL-2B, across different FIDL domains. As shown in Table~\ref{tab:backbone_relative}, although newer general-purpose VLMs may bring gains in some Nature or AIGC scenarios, they show more evident degradation on Document, where local artifacts are crucial. This suggests that unified FIDL performance is not determined only by model generation, parameter scale, or general visual-semantic capability.

We attribute this phenomenon to the way visual information is preserved. InternVL2 adopts a dynamic high-resolution tiling strategy, which better preserves local pixel-level details during visual encoding. In contrast, newer backbones such as InternVL3.5 and Qwen3-VL place more emphasis on inference efficiency, visual token budget control, and semantic aggregation, often introducing stronger compression on the visual side. While such compression is useful for reducing computation and improving semantic modeling in general vision-language tasks, it may dilute or remove weak forensic artifacts in FIDL, such as text-edge changes, local layout anomalies, boundary discontinuities, texture residuals, and compression traces. Therefore, unified FIDL requires a visual backbone that can preserve high-resolution local evidence, rather than simply relying on newer general-purpose VLMs.

\section{Conclusion}
\label{sec:Conclusion}

This paper addresses the growing need for a unified FIDL paradigm in the era of generative foundation models. By proposing \model{}, we demonstrate that a unified approach—integrating powerful vision-language understanding with pixel-level segmentation—is superior to traditional, fragmented detection strategies. Our study yields four key takeaways for the field:

\begin{enumerate}
\item \textbf{Beyond Unconstrained Scaling}:
    We demonstrate that simple data scaling does not follow a linear performance improvement law. Instead, effective unified FIDL models depend on the deliberate balancing of data composition to avoid domain bias and interference.
\item \textbf{Operation-Level Artifacts Awareness}: We reveal that cross-domain transfer is primarily governed by the similarity of underlying manipulation mechanisms (e.g., boundary blending, semantic completion, texture bias) rather than traditional macro-domain labels. Future research should prioritize organizing forensic data based on these operational footprints.
\item \textbf{Multi-Granularity Supervision}: We confirm that while global labels suffice for some tasks, fine-grained pixel-level supervision is essential for tackling the "asymmetry" between fine-grained local manipulations and coarse-grained defense models.
\item \textbf{Original Resolution Artifact Preservation}: 
    We show that unified FIDL requires visual backbones that preserve fine-grained forensic evidence. Newer general-purpose VLMs may improve semantic reasoning or efficiency, but stronger visual token compression can dilute weak artifacts such as text-edge changes, boundary discontinuities, texture residuals, and compression traces. This suggests that FIDL backbone design should prioritize high-resolution local evidence preservation rather than only model scale or general VLM performance.
\end{enumerate}

Experimental results underscore that DeFakerOne provides a robust, generalized, and scalable foundation for detecting modern forgeries. We hope this work provides a strong baseline and serves as a catalyst for moving forensic research toward more holistic, unified, and resilient architectures capable of addressing the challenges posed by next-generation generative models.

\section{Future Work}
\enlargethispage{2\baselineskip}
The successful validation of the DeFakerOne paradigm provides a robust starting point for next-generation digital forensics. However, moving toward a universal, intelligent, and highly generalized forensic infrastructure presents several significant challenges. We outline three critical avenues for future research:

\begin{itemize}
\item \textbf{Scalable Foundation Models and Data Engineering}: Our immediate goal is to evolve DeFakerOne from a standalone model into a foundational infrastructure that supports diverse industrial and academic applications. As demonstrated in this study, the simple accumulation of homogeneous data is insufficient to achieve qualitative leaps in model performance. The key challenge lies in developing high-efficiency data pipelines capable of generating high-fidelity, diverse, and representative forensic data at scale. Furthermore, we aim to explore architectural innovations that enhance model generalization in "in-the-wild" environments, where real-world perturbations and unseen forgery patterns remain highly unpredictable.

\item \textbf{Agentic Paradigms for Expert Knowledge Injection}: We intend to advance the integration of expert knowledge through agentic forensic paradigms, specifically by addressing the dichotomy between outcome injection and process injection. While outcome injection is a generalizable approach, it is increasingly limited by the difficulty of obtaining high-quality expert labels. Process injection—such as the construction of Chain-of-Thought (CoT) data—is a promising but immature solution, currently hampered by content homogenization, high reliance on manual design, and poor knowledge transferability. We posit that the future of forensic intelligence lies in leveraging authentic, low-ambiguity ground-truth data in tandem with agentic tools to move beyond static detection toward sophisticated, evidence-based reasoning.

\item \textbf{A Unified Paradigm for Multi-Modal and Physical-Digital Forensics}:
Beyond the current scope of image-level FIDL, we aim to establish a truly unified forensic paradigm. \\
\textbf{Cross-Modal Integration}: Future work will extend the framework to encompass video and audio authentication, creating a cohesive system for multi-modal integrity verification. \\
\textbf{Physical-Digital Synthesis}: We seek to bridge the gap between digital forgery detection and physical anti-spoofing (e.g., mask, replay, and print attacks). Although these attacks differ in their origin, they share fundamental visual forensic traits, such as local trace anomalies, boundary artifacts, and semantic inconsistencies. By incorporating physical contextual metadata—such as device hardware, sensor noise signatures, and environmental factors—we aim to develop a universal forensic foundation. Designing a framework that bridges the physical and digital divide represents the ultimate frontier in artificial intelligence-driven forensics.
\end{itemize}
\newpage

\bibliographystyle{assets/plainnat}
\bibliography{main}

@String{Computing = "Computing" }

@String{Computer = "{IEEE} Computer" }

@String{Springer = "Springer-Verlag" }

@article{zhang2025ivyfakeunifiedexplainableframework,
  title={Ivy-Fake: A unified explainable framework and benchmark for image and video aigc detection},
  author={Jiang, Changjiang and Dong, Wenhui and Zhang, Zhonghao and Yu, Fengchang and Peng, Wei and Yuan, Xinbin and Bi, Yifei and Zhao, Ming and Zhou, Zian and others},
  journal={arXiv preprint arXiv:2506.00979},
  year={2025}
}

@inproceedings{fakehr1,
  author={Jiang, Changjiang and Sha, Xinkuan and Yu, Fengchang and Liu, Jingjing and Liu, Jian and Fang, Mingqi and Zhang, Chenfeng and Lu, Wei},
  booktitle={ICASSP 2026 - 2026 IEEE International Conference on Acoustics, Speech and Signal Processing (ICASSP)}, 
  title={Fake-HR1: Rethinking Reasoning of Vision Language Model for Synthetic Image Detection}, 
  year={2026},
  pages={10482-10486},
  doi={10.1109/ICASSP55912.2026.11462736}
}

@inproceedings{effort,
  title={Effort: Efficient Orthogonal Modeling for Generalizable AI-Generated Image Detection},
  author={Yan, Zhiyuan and Wang, Jiangming and Wang, Zhendong and Jin, Peng and Zhang, Ke-Yue and Chen, Shen and Yao, Taiping and Ding, Shouhong and Wu, Baoyuan and Yuan, Li},
  booktitle = {ICML},
  year={2024}
}

@inproceedings{zhu2025mesorch,
  title={Mesoscopic insights: Orchestrating multi-scale \& hybrid architecture for image manipulation localization},
  author={Zhu, Xuekang and Ma, Xiaochen and Su, Lei and Jiang, Zhuohang and Du, Bo and Wang, Xiwen and Lei, Zeyu and Feng, Wentao and Pun, Chi-Man and Zhou, Ji-Zhe},
  booktitle={Proceedings of the AAAI Conference on Artificial Intelligence},
  volume={39},
  number={10},
  pages={11022--11030},
  year={2025}
}

@article{forensichub,
  title={Forensichub: A unified benchmark \& codebase for all-domain fake image detection and localization},
  author={Du, Bo and Zhu, Xuekang and Ma, Xiaochen and Qu, Chenfan and Feng, Kaiwen and Yang, Zhe and Pun, Chi-Man and Liu, Jian and Zhou, Ji-Zhe},
  journal={arXiv preprint arXiv:2505.11003},
  year={2025}
}

@inproceedings{fakeclue,
  title={Spot the fake: Large multimodal model-based synthetic image detection with artifact explanation},
  author={Wen, Siwei and Ye, Junyan and Feng, Peilin and Kang, Hengrui and Wen, Zichen and Chen, Yize and Wu, Jiang and Wu, Wenjun and He, Conghui and Li, Weijia},
  booktitle={NeurIPS},
  year={2025}
}

@article{chang2023antifakeprompt,
  title={Antifakeprompt: Prompt-tuned vision-language models are fake image detectors},
  author={Chang, You-Ming and Yeh, Chen and Chiu, Wei-Chen and Yu, Ning},
  journal={arXiv preprint arXiv:2310.17419},
  year={2023}
}

@article{forensicchat,
  title={Seeing before reasoning: A unified framework for generalizable and explainable fake image detection},
  author={Lin, Kaiqing and Yan, Zhiyuan and Chen, Ruoxin and Ye, Junyan and Zhang, Ke-Yue and Zhou, Yue and Jin, Peng and Li, Bin and Yao, Taiping and Ding, Shouhong},
  journal={arXiv preprint arXiv:2509.25502},
  year={2025}
}

@inproceedings{veritas,
  title={Veritas: Generalizable deepfake detection via pattern-aware reasoning},
  author={Tan, Hao and Lan, Jun and Tan, Zichang and Liu, Ajian and Song, Chuanbiao and Shi, Senyuan and Zhu, Huijia and Wang, Weiqiang and Wan, Jun and Lei, Zhen},
  booktitle={ICLR},
  year={2026}
}

@inproceedings{wang2023diffusiondb,
  title={Diffusiondb: A large-scale prompt gallery dataset for text-to-image generative models},
  author={Wang, Zijie J and Montoya, Evan and Munechika, David and Yang, Haoyang and Hoover, Benjamin and Chau, Duen Horng},
  booktitle={Proceedings of the 61st annual meeting of the association for computational linguistics (volume 1: Long papers)},
  pages={893--911},
  year={2023}
}

@inproceedings{cnnspot,
  author={Wang, Sheng-Yu and Wang, Oliver and Zhang, Richard and Owens, Andrew and Efros, Alexei A.},
  booktitle={CVPR}, 
  title={CNN-Generated Images Are Surprisingly Easy to Spot… for Now}, 
  year={2020},
  volume={},
  number={},
  pages={8692-8701},
  doi={10.1109/CVPR42600.2020.00872}}

@article{agentfox,
      title={AgentFoX: LLM Agent-Guided Fusion with eXplainability for AI-Generated Image Detection}, 
      author={Yangxin Yu and Yue Zhou and Bin Li and Kaiqing Lin and Haodong Li and Jiangqun Ni and Bo Cao},
      year={2026},
      journal={arXiv preprint arXiv:2603.23115}
}

@inproceedings{xu2024fakeshield,
  title={Fakeshield: Explainable image forgery detection and localization via multi-modal large language models},
  author={Xu, Zhipei and Zhang, Xuanyu and Li, Runyi and Tang, Zecheng and Huang, Qing and Zhang, Jian},
  booktitle={ICLR},
  year={2025}
}

@article{huang2025unishield,
  title={UniShield: An Adaptive Multi-Agent Framework for Unified Forgery Image Detection and Localization},
  author={Huang, Qing and Xu, Zhipei and Zhang, Xuanyu and Zhang, Jian},
  journal={arXiv preprint arXiv:2510.03161},
  year={2025}
}

@inproceedings{efficientnet,
  title={Rethinking model scaling for convolutional neural networks},
  author={Tan, Mingxing and Le, Q Efficientnet and others},
  booktitle={Proceedings of the International conference on machine learning, Long Beach, CA, USA},
  volume={15},
  year={2019}
}

@article{sabour2017dynamic,
  title={Dynamic routing between capsules},
  author={Sabour, Sara and Frosst, Nicholas and Hinton, Geoffrey E},
  journal={Advances in neural information processing systems},
  volume={30},
  year={2017}
}

@inproceedings{aide,
  title={A sanity check for ai-generated image detection},
  author={Yan, Shilin and Li, Ouxiang and Cai, Jiayin and Hao, Yanbin and Jiang, Xiaolong and Hu, Yao and Xie, Weidi},
  booktitle={ICLR},
  year={2025}
}

@INPROCEEDINGS{dire,
  author={Wang, Zhendong and Bao, Jianmin and Zhou, Wengang and Wang, Weilun and Hu, Hezhen and Chen, Hong and Li, Houqiang},
  booktitle={ICCV}, 
  title={DIRE for Diffusion-Generated Image Detection}, 
  year={2023},
  pages={22388-22398},
  doi={10.1109/ICCV51070.2023.02051}}

@InProceedings{forensic-moe,
    author    = {Fang, Mingqi and Li, Ziguang and Yu, Lingyun and Yang, Quanwei and Xie, Hongtao and Zhang, Yongdong},
    title     = {Forensic-MoE: Exploring Comprehensive Synthetic Image Detection Traces with Mixture of Experts},
    booktitle = {Proceedings of the IEEE/CVF International Conference on Computer Vision (ICCV)},
    month     = {October},
    year      = {2025},
    pages     = {17772-17782}
}

@inproceedings{CASIA_2013,
  address      = {Beijing, China},
  title        = {CASIA Image Tampering Detection Evaluation Database},
  isbn         = {978-1-4799-1043-4},
  url          = {http://ieeexplore.ieee.org/document/6625374/},
  doi          = {10.1109/ChinaSIP.2013.6625374},
  abstractnote = {Image forensics has now raised the anxiety of justice as increasing cases of abusing tampered images in newspapers and court for evidence are reported recently. With the goal of verifying image content authenticity, passive-blind image tampering detection is called for. More realistic open benchmark databases are also needed to assist the techniques. Recently, we collect a natural color image database with realistic tampering operations. The database is made publicly available for researchers to compare and evaluate their proposed tampering detection techniques. We call this database CASIA Image Tampering Detection Evaluation Database. We describe the purpose, the design criterion, the organization and self-evaluation of this database in this paper.},
  booktitle    = {2013 IEEE China Summit and International Conference on Signal and Information Processing},
  publisher    = {IEEE},
  author       = {Dong, Jing and Wang, Wei and Tan, Tieniu},
  year         = {2013},
  month        = {Jul},
  pages        = {422–426},
  language     = {en}
}

@inproceedings{Columbia_2006,
  address      = {Toronto, ON, Canada},
  title        = {Detecting Image Splicing using Geometry Invariants and Camera Characteristics Consistency},
  isbn         = {978-1-4244-0367-7},
  url          = {http://ieeexplore.ieee.org/document/4036658/},
  doi          = {10.1109/ICME.2006.262447},
  abstractnote = {Recent advances in computer technology have made digital image tampering more and more common. In this paper, we propose an authentic vs. spliced image classiﬁcation method making use of geometry invariants in a semi-automatic manner. For a given image, we identify suspicious splicing areas, compute the geometry invariants from the pixels within each region, and then estimate the camera response function (CRF) from these geometry invariants. The cross-ﬁtting errors are fed into a statistical classiﬁer. Experiments show a very promising accuracy, 87\%, over a large data set of 363 natural and spliced images. To the best of our knowledge, this is the ﬁrst work detecting image splicing by verifying camera characteristic consistency from a single-channel image.},
  booktitle    = {2006 IEEE International Conference on Multimedia and Expo},
  publisher    = {IEEE},
  author       = {Hsu, Yu-feng and Chang, Shih-fu},
  year         = {2006},
  month        = {Jul},
  pages        = {549–552},
  language     = {en}
}

@inproceedings{Coverage_2016,
  address   = {Phoenix, AZ, USA},
  title     = {COVERAGE — A novel database for copy-move forgery detection},
  isbn      = {978-1-4673-9961-6},
  url       = {http://ieeexplore.ieee.org/document/7532339/},
  doi       = {10.1109/ICIP.2016.7532339},
  booktitle = {2016 IEEE International Conference on Image Processing (ICIP)},
  publisher = {IEEE},
  author    = {Wen, Bihan and Zhu, Ye and Subramanian, Ramanathan and Ng, Tian-Tsong and Shen, Xuanjing and Winkler, Stefan},
  year      = {2016},
  month     = {Sep},
  pages     = {161–165},
  language  = {en}
}

@inproceedings{defacto_2019,
  address      = {A Coruna, Spain},
  title        = {DEFACTO: Image and Face Manipulation Dataset},
  isbn         = {978-90-827970-3-9},
  url          = {https://ieeexplore.ieee.org/document/8903181/},
  doi          = {10.23919/EUSIPCO.2019.8903181},
  abstractnote = {This paper presents a novel dataset for image and face manipulation detection and localization called DEFACTO. The dataset was automatically generated using Microsoft common object in context database (MSCOCO) to produce semantically meaningful forgeries. Four categories of forgeries have been generated. Splicing forgeries which consist of inserting an external element into an image, copy-move forgeries where an element within an image is duplicated, object removal forgeries where objects are removed from images and lastly morphing where two images are warped and blended together. Over 200000 images have been generated and each image is accompanied by several annotations allowing precise localization of the forgery and information about the tampering process.},
  booktitle    = {2019 27th European Signal Processing Conference (EUSIPCO)},
  publisher    = {IEEE},
  author       = {Mahfoudi, Gael and Tajini, Badr and Retraint, Florent and Morain-Nicolier, Frederic and Dugelay, Jean Luc and Pic, Marc},
  year         = {2019},
  month        = {Sep},
  pages        = {1–5},
  language     = {en}
}

@inbook{mscoco_2014,
  address      = {Cham},
  series       = {Lecture Notes in Computer Science},
  title        = {Microsoft COCO: Common Objects in Context},
  volume       = {8693},
  isbn         = {978-3-319-10601-4},
  url          = {http://link.springer.com/10.1007/978-3-319-10602-1_48},
  doi          = {10.1007/978-3-319-10602-1_48},
  abstractnote = {We present a new dataset with the goal of advancing the state-of-the-art in object recognition by placing the question of object recognition in the context of the broader question of scene understanding. This is achieved by gathering images of complex everyday scenes containing common objects in their natural context. Objects are labeled using per-instance segmentations to aid in precise object localization. Our dataset contains photos of 91 objects types that would be easily recognizable by a 4 year old. With a total of 2.5 million labeled instances in 328k images, the creation of our dataset drew upon extensive crowd worker involvement via novel user interfaces for category detection, instance spotting and instance segmentation. We present a detailed statistical analysis of the dataset in comparison to PASCAL, ImageNet, and SUN. Finally, we provide baseline performance analysis for bounding box and segmentation detection results using a Deformable Parts Model.},
  booktitle    = {Computer Vision – ECCV 2014},
  publisher    = {Springer International Publishing},
  author       = {Lin, Tsung-Yi and Maire, Michael and Belongie, Serge and Hays, James and Perona, Pietro and Ramanan, Deva and Dollár, Piotr and Zitnick, C. Lawrence},
  editor       = {Fleet, David and Pajdla, Tomas and Schiele, Bernt and Tuytelaars, Tinne},
  year         = {2014},
  pages        = {740–755},
  collection   = {Lecture Notes in Computer Science},
  language     = {en}
}

@inproceedings{NIST16_2019,
  address   = {Waikoloa Village, HI, USA},
  title     = {MFC Datasets: Large-Scale Benchmark Datasets for Media Forensic Challenge Evaluation},
  isbn      = {978-1-72811-392-0},
  url       = {https://ieeexplore.ieee.org/document/8638296/},
  doi       = {10.1109/WACVW.2019.00018},
  booktitle = {2019 IEEE Winter Applications of Computer Vision Workshops (WACVW)},
  publisher = {IEEE},
  author    = {Guan, Haiying and Kozak, Mark and Robertson, Eric and Lee, Yooyoung and Yates, Amy N. and Delgado, Andrew and Zhou, Daniel and Kheyrkhah, Timothee and Smith, Jeff and Fiscus, Jonathan},
  year      = {2019},
  month     = {Jan},
  pages     = {63–72}
}

@inproceedings{Resnet_2016,
  address      = {Las Vegas, NV, USA},
  title        = {Deep Residual Learning for Image Recognition},
  isbn         = {978-1-4673-8851-1},
  url          = {http://ieeexplore.ieee.org/document/7780459/},
  doi          = {10.1109/CVPR.2016.90},
  abstractnote = {Deeper neural networks are more difﬁcult to train. We present a residual learning framework to ease the training of networks that are substantially deeper than those used previously. We explicitly reformulate the layers as learning residual functions with reference to the layer inputs, instead of learning unreferenced functions. We provide comprehensive empirical evidence showing that these residual networks are easier to optimize, and can gain accuracy from considerably increased depth. On the ImageNet dataset we evaluate residual nets with a depth of up to 152 layers—8× deeper than VGG nets [40] but still having lower complexity. An ensemble of these residual nets achieves 3.57\% error on the ImageNet test set. This result won the 1st place on the ILSVRC 2015 classiﬁcation task. We also present analysis on CIFAR-10 with 100 and 1000 layers.},
  booktitle    = {2016 IEEE Conference on Computer Vision and Pattern Recognition (CVPR)},
  publisher    = {IEEE},
  author       = {He, Kaiming and Zhang, Xiangyu and Ren, Shaoqing and Sun, Jian},
  year         = {2016},
  month        = {Jun},
  pages        = {770–778},
  language     = {en}
}

@article{SegFormer_2021,
  title={SegFormer: Simple and efficient design for semantic segmentation with transformers},
  author={Xie, Enze and Wang, Wenhai and Yu, Zhiding and Anandkumar, Anima and Alvarez, Jose M and Luo, Ping},
  journal={Advances in neural information processing systems},
  volume={34},
  pages={12077--12090},
  year={2021}
}

@article{mvsspp_2022,
  title        = {MVSS-Net: Multi-View Multi-Scale Supervised Networks for Image Manipulation Detection},
  issn         = {0162-8828, 2160-9292, 1939-3539},
  doi          = {10.1109/TPAMI.2022.3180556},
  abstractnote = {As manipulating images by copy-move, splicing and/or inpainting may lead to misinterpretation of the visual content, detecting these sorts of manipulations is crucial for media forensics. Given the variety of possible attacks on the content, devising a generic method is nontrivial. Current deep learning based methods are promising when training and test data are well aligned, but perform poorly on independent tests. Moreover, due to the absence of authentic test images, their image-level detection speciﬁcity is in doubt. The key question is how to design and train a deep neural network capable of learning generalizable features sensitive to manipulations in novel data, whilst speciﬁc to prevent false alarms on the authentic. We propose multi-view feature learning to jointly exploit tampering boundary artifacts and the noise view of the input image. As both clues are meant to be semantic-agnostic, the learned features are thus generalizable. For effectively learning from authentic images, we train with multi-scale (pixel / edge / image) supervision. We term the new network MVSS-Net and its enhanced version MVSS-Net++. Experiments are conducted in both within-dataset and cross-dataset scenarios, showing that MVSS-Net++ performs the best, and exhibits better robustness against JPEG compression, Gaussian blur and screenshot based image re-capturing.},
  journal      = {IEEE Transactions on Pattern Analysis and Machine Intelligence},
  author       = {Dong, Chengbo and Chen, Xinru and Hu, Ruohan and Cao, Juan and Li, Xirong},
  year         = {2022},
  pages        = {1–14},
  language     = {en}
}

@article{sa2va,
  title={Sa2va: Marrying sam2 with llava for dense grounded understanding of images and videos},
  author={Yuan, Haobo and Li, Xiangtai and Zhang, Tao and Sun, Yueyi and Huang, Zilong and Xu, Shilin and Ji, Shunping and Tong, Yunhai and Qi, Lu and Feng, Jiashi and others},
  journal={arXiv preprint arXiv:2501.04001},
  year={2025}
}

@inproceedings{sam2,
  title={Sam 2: Segment anything in images and videos},
  author={Ravi, Nikhila and Gabeur, Valentin and Hu, Yuan-Ting and Hu, Ronghang and Ryali, Chaitanya and Ma, Tengyu and Khedr, Haitham and R{\"a}dle, Roman and Rolland, Chloe and Gustafson, Laura and others},
  booktitle={International Conference on Learning Representations},
  volume={2025},
  pages={28085--28128},
  year={2025}
}

@inproceedings{trufor2023,
  title={TruFor: Leveraging all-round clues for trustworthy image forgery detection and localization},
  author={Guillaro, Fabrizio and Cozzolino, Davide and Sud, Avneesh and Dufour, Nicholas and Verdoliva, Luisa},
  booktitle={Proceedings of the IEEE/CVF Conference on Computer Vision and Pattern Recognition},
  pages={20606--20615},
  year={2023}
}

@article{CAT-Net2022,
  title={Learning JPEG compression artifacts for image manipulation detection and localization},
  author={Kwon, Myung-Joon and Nam, Seung-Hun and Yu, In-Jae and Lee, Heung-Kyu and Kim, Changick},
  journal={International Journal of Computer Vision},
  volume={130},
  number={8},
  pages={1875--1895},
  year={2022},
  publisher={Springer}
}

@inproceedings{IMD20_2020, address={Snowmass Village, CO, USA}, title={IMD2020: A Large-Scale Annotated Dataset Tailored for Detecting Manipulated Images}, ISBN={978-1-72817-162-3}, url={https://ieeexplore.ieee.org/document/9096940/}, DOI={10.1109/WACVW50321.2020.9096940}, abstractNote={Witnessing impressive results of deep nets in a number of computer vision problems, the image forensic community has begun to utilize them in the challenging domain of detecting manipulated visual content. One of the obstacles to replicate the success of deep nets here is absence of diverse datasets tailored for training and testing of image forensic methods. Such datasets need to be designed to capture wide and complex types of systematic noise and intrinsic artifacts of images in order to avoid overﬁtting of learning methods to just a narrow set of camera types or types of manipulations. These artifacts are brought into visual content by various components of the image acquisition process as well as the manipulating process.}, booktitle={2020 IEEE Winter Applications of Computer Vision Workshops (WACVW)}, publisher={IEEE}, author={Novozamsky, Adam and Mahdian, Babak and Saic, Stanislav}, year={2020}, month=mar, pages={71–80}, language={en} }

@inproceedings{NCL_IML_2023,
  title={Pre-training-free Image Manipulation Localization through Non-Mutually Exclusive Contrastive Learning},
  author={Zhou, Jizhe and Ma, Xiaochen and Du, Xia and Alhammadi, Ahmed Y and Feng, Wentao},
  booktitle={Proceedings of the IEEE/CVF International Conference on Computer Vision},
  pages={22346--22356},
  year={2023}
}

@article{ma2023iml,
  title={Iml-vit: Image manipulation localization by vision transformer},
  author={Ma, Xiaochen and Du, Bo and Liu, Xianggen and Hammadi, Ahmed Y Al and Zhou, Jizhe},
  journal={arXiv preprint arXiv:2307.14863},
  year={2023}
}

@inproceedings{diffusion_detection,
  title={On the detection of synthetic images generated by diffusion models},
  author={Corvi, Riccardo and Cozzolino, Davide and Zingarini, Giada and Poggi, Giovanni and Nagano, Koki and Verdoliva, Luisa},
  booktitle={ICASSP 2023-2023 IEEE International Conference on Acoustics, Speech and Signal Processing (ICASSP)},
  pages={1--5},
  year={2023},
  organization={IEEE}
}

@article{ma2024imdl,
  title={Imdl-benco: A comprehensive benchmark and codebase for image manipulation detection \& localization},
  author={Ma, Xiaochen and Zhu, Xuekang and Su, Lei and Du, Bo and Jiang, Zhuohang and Tong, Bingkui and Lei, Zeyu and Yang, Xinyu and Pun, Chi-Man and Lv, Jiancheng and others},
  journal={Advances in Neural Information Processing Systems},
  volume={37},
  pages={134591--134613},
  year={2024}
}

@inproceedings{jia2023autosplice,
  title={Autosplice: A text-prompt manipulated image dataset for media forensics},
  author={Jia, Shan and Huang, Mingzhen and Zhou, Zhou and Ju, Yan and Cai, Jialing and Lyu, Siwei},
  booktitle={Proceedings of the IEEE/CVF conference on computer vision and pattern recognition},
  pages={893--903},
  year={2023}
}

@article{zhu2023genimage,
  title={Genimage: A million-scale benchmark for detecting ai-generated image},
  author={Zhu, Mingjian and Chen, Hanting and Yan, Qiangyu and Huang, Xudong and Lin, Guanyu and Li, Wei and Tu, Zhijun and Hu, Hailin and Hu, Jie and Wang, Yunhe},
  journal={Advances in Neural Information Processing Systems},
  volume={36},
  pages={77771--77782},
  year={2023}
}

@inproceedings{dualnet,
  title={Ai-generated image detection using a cross-attention enhanced dual-stream network},
  author={Xi, Ziyi and Huang, Wenmin and Wei, Kangkang and Luo, Weiqi and Zheng, Peijia},
  booktitle={2023 Asia Pacific Signal and Information Processing Association Annual Summit and Conference (APSIPA ASC)},
  pages={1463--1470},
  year={2023},
  organization={IEEE}
}

@inproceedings{evalgen,
  title={Dual Data Alignment Makes {AI}-Generated Image Detector Easier Generalizable},
  author={Ruoxin Chen and Junwei Xi and Zhiyuan Yan and Ke-Yue Zhang and Shuang Wu and Jingyi Xie and Xu Chen and Lei Xu and Isabel Guan and Taiping Yao and Shouhong Ding},
  booktitle={The Thirty-ninth Annual Conference on Neural Information Processing Systems},
  year={2025},
  url={https://openreview.net/forum?id=C39ShJwtD5}
}

@inproceedings{univfd,
  title={Towards universal fake image detectors that generalize across generative models},
  author={Ojha, Utkarsh and Li, Yuheng and Lee, Yong Jae},
  booktitle={Proceedings of the IEEE/CVF Conference on Computer Vision and Pattern Recognition},
  pages={24480--24489},
  year={2023}
}

@inproceedings{li2020celeb,
author = {Li, Yuezun and Yang, Xin and Sun, Pu and Qi, Honggang and Lyu, Siwei},
title = {Celeb-DF: A Large-Scale Challenging Dataset for DeepFake Forensics},
booktitle = {Proceedings of the IEEE/CVF Conference on Computer Vision and Pattern Recognition (CVPR)},
month = {June},
year = {2020}
}

@article{dolhansky2020dfd,
  title={The deepfake detection challenge (dfdc) dataset},
  author={Dolhansky, Brian and Bitton, Joanna and Pflaum, Ben and Lu, Jikuo and Howes, Russ and Wang, Menglin and Ferrer, Cristian Canton},
  journal={arXiv preprint arXiv:2006.07397},
  year={2020}
}

@article{dolhansky2019deepfake,
  title={The deepfake detection challenge (dfdc) preview dataset},
  author={Dolhansky, Brian and Howes, Russ and Pflaum, Ben and Baram, Nicole and Ferrer, Cristian Canton},
  journal={arXiv preprint arXiv:1910.08854},
  year={2019}
}

@inproceedings{rossler2019faceforensics++,
  title={Faceforensics++: Learning to detect manipulated facial images},
  author={Rossler, Andreas and Cozzolino, Davide and Verdoliva, Luisa and Riess, Christian and Thies, Justus and Nie{\ss}ner, Matthias},
  booktitle={Proceedings of the IEEE/CVF international conference on computer vision},
  pages={1--11},
  year={2019}
}

@article{li2019faceshifter,
  title={Faceshifter: Towards high fidelity and occlusion aware face swapping},
  author={Li, Lingzhi and Bao, Jianmin and Yang, Hao and Chen, Dong and Wen, Fang},
  journal={arXiv preprint arXiv:1912.13457},
  year={2019}
}

@article{wang2022tsroie,
  title={Tampered text detection via rgb and frequency relationship modeling},
  author={Wang, Yuxin and Zhang, Boqiang and Xie, Hongtao and Zhang, Yongdong},
  journal={Chinese Journal of Network and Information Security},
  volume={8},
  number={3},
  pages={29--40},
  year={2022}
}

@inproceedings{wang2022tpic,
  title={Detecting tampered scene text in the wild},
  author={Wang, Yuxin and Xie, Hongtao and Xing, Mengting and Wang, Jing and Zhu, Shenggao and Zhang, Yongdong},
  booktitle={European Conference on Computer Vision},
  pages={215--232},
  year={2022},
  organization={Springer}
}

@article{luo2025rtm,
  title={Toward real text manipulation detection: New dataset and new solution},
  author={Luo, Dongliang and Liu, Yuliang and Yang, Rui and Liu, Xianjin and Zeng, Jishen and Zhou, Yu and Bai, Xiang},
  journal={Pattern Recognition},
  volume={157},
  pages={110828},
  year={2025},
  publisher={Elsevier}
}

@inproceedings{chen2024ffdn,
  title={Enhancing Tampered Text Detection Through Frequency Feature Fusion and Decomposition},
  author={Chen, Zhongxi and Chen, Shen and Yao, Taiping and Sun, Ke and Ding, Shouhong and Lin, Xianming and Cao, Liujuan and Ji, Rongrong},
  booktitle={European Conference on Computer Vision},
  pages={200--217},
  year={2024},
  organization={Springer}
}

@article{song2025caftb,
  title={Cross-attention based two-branch networks for document image forgery localization in the Metaverse},
  author={Song, Yalin and Jiang, Wenbin and Chai, Xiuli and Gan, Zhihua and Zhou, Mengyuan and Chen, Lei},
  journal={ACM Transactions on Multimedia Computing, Communications and Applications},
  volume={21},
  number={2},
  pages={1--24},
  year={2025},
  publisher={ACM New York, NY}
}

@article{dong2024tifdm,
  title={Robust text image tampering localization via forgery traces enhancement and multiscale attention},
  author={Dong, Li and Liang, Weipeng and Wang, Rangding},
  journal={IEEE Transactions on Consumer Electronics},
  year={2024},
  publisher={IEEE}
}

@inproceedings{nguyen2019capsule,
  title={Capsule-forensics: Using capsule networks to detect forged images and videos},
  author={Nguyen, Huy H and Yamagishi, Junichi and Echizen, Isao},
  booktitle={ICASSP 2019-2019 IEEE international conference on acoustics, speech and signal processing (ICASSP)},
  pages={2307--2311},
  year={2019},
  organization={IEEE}
}

@inproceedings{cao2022recce,
  title={End-to-end reconstruction-classification learning for face forgery detection},
  author={Cao, Junyi and Ma, Chao and Yao, Taiping and Chen, Shen and Ding, Shouhong and Yang, Xiaokang},
  booktitle={Proceedings of the IEEE/CVF conference on computer vision and pattern recognition},
  pages={4113--4122},
  year={2022}
}

@inproceedings{liu2021spsl,
  title={Spatial-phase shallow learning: rethinking face forgery detection in frequency domain},
  author={Liu, Honggu and Li, Xiaodan and Zhou, Wenbo and Chen, Yuefeng and He, Yuan and Xue, Hui and Zhang, Weiming and Yu, Nenghai},
  booktitle={Proceedings of the IEEE/CVF conference on computer vision and pattern recognition},
  pages={772--781},
  year={2021}
}

@inproceedings{shiohara2022sbi,
  title={Detecting deepfakes with self-blended images},
  author={Shiohara, Kaede and Yamasaki, Toshihiko},
  booktitle={Proceedings of the IEEE/CVF conference on computer vision and pattern recognition},
  pages={18720--18729},
  year={2022}
}

@inproceedings{opensdi,
  title={Opensdi: Spotting diffusion-generated images in the open world},
  author={Wang, Yabin and Huang, Zhiwu and Hong, Xiaopeng},
  booktitle={Proceedings of the IEEE/CVF Conference on Computer Vision and Pattern Recognition},
  pages={4291--4301},
  year={2025}
}

@InProceedings{CommunityForensics,
    author    = {Park, Jeongsoo and Owens, Andrew},
    title     = {Community Forensics: Using Thousands of Generators to Train Fake Image Detectors},
    booktitle = {Proceedings of the Computer Vision and Pattern Recognition Conference (CVPR)},
    month     = {June},
    year      = {2025},
    pages     = {8245-8257}
}

@InProceedings{bfree,
    author    = {Guillaro, Fabrizio and Zingarini, Giada and Usman, Ben and Sud, Avneesh and Cozzolino, Davide and Verdoliva, Luisa},
    title     = {A Bias-Free Training Paradigm for More General AI-generated Image Detection},
    booktitle = {Proceedings of the Computer Vision and Pattern Recognition Conference (CVPR)},
    month     = {June},
    year      = {2025},
    pages     = {18685-18694}
}

@InProceedings{drct,
  title = 	 {{DRCT}: Diffusion Reconstruction Contrastive Training towards Universal Detection of Diffusion Generated Images},
  author =       {Chen, Baoying and Zeng, Jishen and Yang, Jianquan and Yang, Rui},
  booktitle = 	 {Proceedings of the 41st International Conference on Machine Learning},
  pages = 	 {7621--7639},
  year = 	 {2024},
  editor = 	 {Salakhutdinov, Ruslan and Kolter, Zico and Heller, Katherine and Weller, Adrian and Oliver, Nuria and Scarlett, Jonathan and Berkenkamp, Felix},
  volume = 	 {235},
  series = 	 {Proceedings of Machine Learning Research},
  month = 	 {21--27 Jul},
  publisher =    {PMLR},
  url = 	 {https://proceedings.mlr.press/v235/chen24ay.html}
}

@inproceedings{SynthWildx,
  author={Davide Cozzolino and Giovanni Poggi and Riccardo Corvi and Matthias Nießner and Luisa Verdoliva},
  title={Raising the Bar of AI-generated Image Detection with CLIP}, 
  booktitle={IEEE/CVF Conference on Computer Vision and Pattern Recognition Workshops (CVPRW)},
  year={2024},
}

@ARTICLE{dso-1,
  author={de Carvalho, Tiago José and Riess, Christian and Angelopoulou, Elli and Pedrini, Hélio and de Rezende Rocha, Anderson},
  journal={IEEE Transactions on Information Forensics and Security}, 
  title={Exposing Digital Image Forgeries by Illumination Color Classification}, 
  year={2013},
  volume={8},
  number={7},
  pages={1182-1194},
  doi={10.1109/TIFS.2013.2265677}}

@misc{dfd2019,
  author       = {{Google AI Blog}},
  title        = {Contributing Data to Deepfake Detection},
  howpublished = {\url{https://ai.googleblog.com/2019/09/contributing-data-to-deepfake-detection.html}},
  note         = {Accessed 2025-04-25},
  year         = {2019}
}

@inproceedings{Swin_2021,
  address      = {Montreal, QC, Canada},
  title        = {Swin Transformer: Hierarchical Vision Transformer using Shifted Windows},
  isbn         = {978-1-66542-812-5},
  url          = {https://ieeexplore.ieee.org/document/9710580/},
  doi          = {10.1109/ICCV48922.2021.00986},
  abstractnote = {This paper presents a new vision Transformer, called Swin Transformer, that capably serves as a general-purpose backbone for computer vision. Challenges in adapting Transformer from language to vision arise from differences between the two domains, such as large variations in the scale of visual entities and the high resolution of pixels in images compared to words in text. To address these differences, we propose a hierarchical Transformer whose representation is computed with Shifted windows. The shifted windowing scheme brings greater efﬁciency by limiting self-attention computation to non-overlapping local windows while also allowing for cross-window connection. This hierarchical architecture has the ﬂexibility to model at various scales and has linear computational complexity with respect to image size. These qualities of Swin Transformer make it compatible with a broad range of vision tasks, including image classiﬁcation (87.3 top-1 accuracy on ImageNet-1K) and dense prediction tasks such as object detection (58.7 box AP and 51.1 mask AP on COCO testdev) and semantic segmentation (53.5 mIoU on ADE20K val). Its performance surpasses the previous state-of-theart by a large margin of +2.7 box AP and +2.6 mask AP on COCO, and +3.2 mIoU on ADE20K, demonstrating the potential of Transformer-based models as vision backbones. The hierarchical design and the shifted window approach also prove beneﬁcial for all-MLP architectures. The code and models are publicly available at https://github. com/microsoft/Swin-Transformer.},
  booktitle    = {2021 IEEE/CVF International Conference on Computer Vision (ICCV)},
  publisher    = {IEEE},
  author       = {Liu, Ze and Lin, Yutong and Cao, Yue and Hu, Han and Wei, Yixuan and Zhang, Zheng and Lin, Stephen and Guo, Baining},
  year         = {2021},
  month        = {Oct},
  pages        = {9992–10002},
  language     = {en}
}

@inproceedings{HiFi-Net2023,
  title={Hierarchical fine-grained image forgery detection and localization},
  author={Guo, Xiao and Liu, Xiaohong and Ren, Zhiyuan and Grosz, Steven and Masi, Iacopo and Liu, Xiaoming},
  booktitle={Proceedings of the IEEE/CVF Conference on Computer Vision and Pattern Recognition},
  pages={3155--3165},
  year={2023}
}

@article{li2018uadfv,
  title={Exposing deepfake videos by detecting face warping artifacts},
  author={Li, Yuezun and Lyu, Siwei},
  journal={arXiv preprint arXiv:1811.00656},
  year={2018}
}

@inproceedings{sparseViT_2025,
  title={Can we get rid of handcrafted feature extractors? sparsevit: Nonsemantics-centered, parameter-efficient image manipulation localization through spare-coding transformer},
  author={Su, Lei and Ma, Xiaochen and Zhu, Xuekang and Niu, Chaoqun and Lei, Zeyu and Zhou, Ji-Zhe},
  booktitle={Proceedings of the AAAI Conference on Artificial Intelligence},
  volume={39},
  number={7},
  pages={7024--7032},
  year={2025}
}

@inproceedings{sia,
  title={An information theoretic approach for attention-driven face forgery detection},
  author={Sun, Ke and Liu, Hong and Yao, Taiping and Sun, Xiaoshuai and Chen, Shen and Ding, Shouhong and Ji, Rongrong},
  booktitle={European conference on computer vision},
  pages={111--127},
  year={2022},
  organization={Springer}
}

@inproceedings{fatformer,
  title={Forgery-aware adaptive transformer for generalizable synthetic image detection},
  author={Liu, Huan and Tan, Zichang and Tan, Chuangchuang and Wei, Yunchao and Wang, Jingdong and Zhao, Yao},
  booktitle={Proceedings of the IEEE/CVF Conference on Computer Vision and Pattern Recognition},
  pages={10770--10780},
  year={2024}
}

@inproceedings{co-spy,
  title={CO-SPY: Combining Semantic and Pixel Features to Detect Synthetic Images by AI},
  author={Cheng, Siyuan and Lyu, Lingjuan and Wang, Zhenting and Zhang, Xiangyu and Sehwag, Vikash},
  booktitle={Proceedings of the Computer Vision and Pattern Recognition Conference},
  pages={13455--13465},
  year={2025}
}

@article{yan2024df40,
  title={Df40: Toward next-generation deepfake detection},
  author={Yan, Zhiyuan and Yao, Taiping and Chen, Shen and Zhao, Yandan and Fu, Xinghe and Zhu, Junwei and Luo, Donghao and Wang, Chengjie and Ding, Shouhong and Wu, Yunsheng and others},
  journal={Advances in Neural Information Processing Systems},
  volume={37},
  pages={29387--29434},
  year={2024}
}

@article{zhu2025does,
  title={Does the Manipulation Process Matter? RITA: Reasoning Composite Image Manipulations via Reversely-Ordered Incremental-Transition Autoregression},
  author={Zhu, Xuekang and Zhou, Ji-Zhe and Feng, Kaiwen and Qu, Chenfan and Wang, Yunfei and Zhou, Liting and Liu, Jian},
  journal={arXiv preprint arXiv:2509.20006},
  year={2025}
}

@inproceedings{sida,
  author       = {Zhenglin Huang and Jinwei Hu and Xiangtai Li and Yiwei He and Xingyu Zhao and Bei Peng and Baoyuan Wu and Xiaowei Huang and Guangliang Cheng},
  title        = {{SIDA:} Social Media Image Deepfake Detection, Localization and Explanation
                  with Large Multimodal Model},
  booktitle    = {{IEEE/CVF} Conference on Computer Vision and Pattern Recognition(CVPR) 2025},
  year         = {2025},
}

@inproceedings{chen2024internvl,
    title={Internvl: Scaling up vision foundation models and aligning for generic visual-linguistic tasks},
    author={Chen, Zhe and Wu, Jiannan and Wang, Wenhai and Su, Weijie and Chen, Guo and Xing, Sen and Zhong, Muyan and Zhang, Qinglong and Zhu, Xizhou and Lu, Lewei and others},
    booktitle={Proceedings of the IEEE/CVF Conference on Computer Vision and Pattern Recognition},
    pages={24185--24198},
    year={2024}
  }

@inproceedings{guo2025rethinking,
  title={Rethinking vision-language model in face forensics: Multi-modal interpretable forged face detector},
  author={Guo, Xiao and Song, Xiufeng and Zhang, Yue and Liu, Xiaohong and Liu, Xiaoming},
  booktitle={Proceedings of the Computer Vision and Pattern Recognition Conference},
  pages={105--116},
  year={2025}
}

@article{huang2025thinkfake,
  title={Thinkfake: Reasoning in multimodal large language models for ai-generated image detection},
  author={Huang, Tai-Ming and Lin, Wei-Tung and Hua, Kai-Lung and Cheng, Wen-Huang and Yamagishi, Junichi and Chen, Jun-Cheng},
  journal={arXiv preprint arXiv:2509.19841},
  year={2025}
}

@article{liu2024forgerygpt,
  title={Forgerygpt: Multimodal large language model for explainable image forgery detection and localization},
  author={Liu, Jiawei and Zhang, Fanrui and Zhu, Jiaying and Sun, Esther and Zhang, Qiang and Zha, Zheng-Jun},
  journal={arXiv preprint arXiv:2410.10238},
  year={2024}
}

@article{xia2025mirage,
  title={MIRAGE: Towards AI-Generated Image Detection in the Wild},
  author={Xia, Cheng and Lin, Manxi and Tan, Jiexiang and Du, Xiaoxiong and Qiu, Yang and Zheng, Junjun and Kong, Xiangheng and Jiang, Yuning and Zheng, Bo},
  journal={arXiv preprint arXiv:2508.13223},
  year={2025}
}

@inproceedings{kang2025legion,
  title={Legion: Learning to ground and explain for synthetic image detection},
  author={Kang, Hengrui and Wen, Siwei and Wen, Zichen and Ye, Junyan and Li, Weijia and Feng, Peilin and Zhou, Baichuan and Wang, Bin and Lin, Dahua and Zhang, Linfeng and others},
  booktitle={Proceedings of the IEEE/CVF International Conference on Computer Vision},
  pages={18937--18947},
  year={2025}
}

@article{x2-dfd,
  title={X2-dfd: A framework for explainable and extendable deepfake detection},
  author={Chen, Yize and Yan, Zhiyuan and Cheng, Guangliang and Zhao, Kangran and Lyu, Siwei and Wu, Baoyuan},
  journal={arXiv preprint arXiv:2410.06126},
  year={2024}
}

@article{fakescope,
  title={FakeScope: Large multimodal expert model for transparent AI-generated image forensics},
  author={Li, Yixuan and Tian, Yu and Huang, Yipo and Lu, Wei and Wang, Shiqi and Lin, Weisi and Rocha, Anderson},
  journal={arXiv preprint arXiv:2503.24267},
  year={2025}
}

@article{zhu2026evoguard,
  title={EvoGuard: An Extensible Agentic RL-based Framework for Practical and Evolving AI-Generated Image Detection},
  author={Zhu, Chenyang and Wang, Maorong and Liu, Jun and Chang, Ching-Chun and Echizen, Isao},
  journal={arXiv preprint arXiv:2603.17343},
  year={2026}
}

@article{chen2025task,
  title={Task-Model Alignment: A Simple Path to Generalizable AI-Generated Image Detection},
  author={Chen, Ruoxin and Gao, Jiahui and Lin, Kaiqing and Zhang, Keyue and Zhao, Yandan and Guan, Isabel and Yao, Taiping and Ding, Shouhong},
  journal={arXiv preprint arXiv:2512.06746},
  year={2025}
}

@inproceedings{zi2020wilddeepfake,
  title={Wilddeepfake: A challenging real-world dataset for deepfake detection},
  author={Zi, Bojia and Chang, Minghao and Chen, Jingjing and Ma, Xingjun and Jiang, Yu-Gang},
  booktitle={Proceedings of the 28th ACM International Conference on Multimedia},
  pages={2382--2390},
  year={2020}
}

@inproceedings{fakeinversion,
  title={Fakeinversion: Learning to detect images from unseen text-to-image models by inverting stable diffusion},
  author={Cazenavette, George and Sud, Avneesh and Leung, Thomas and Usman, Ben},
  booktitle={Proceedings of the IEEE/CVF Conference on Computer Vision and Pattern Recognition},
  pages={10759--10769},
  year={2024}
}

@article{scaledf,
  title={Scaling Laws for Deepfake Detection},
  author={Wang, Wenhao and Cai, Longqi and Xiao, Taihong and Wang, Yuxiao and Yang, Ming-Hsuan},
  journal={arXiv preprint arXiv:2510.16320},
  year={2025}
}

@inproceedings{miao2025mffi,
  title={MFFI: Multi-dimensional face forgery image dataset for real-world scenarios},
  author={Miao, Changtao and Zhang, Yi and Luo, Man and Feng, Weiwei and Zheng, Kaiyuan and Chu, Qi and Gong, Tao and Li, Jianshu and Diao, Yunfeng and Zhou, Wei and others},
  booktitle={Proceedings of the 33rd ACM International Conference on Multimedia},
  pages={13235--13242},
  year={2025}
}

@misc{rific,
  title        = {AI Identity Verification: Financial Certificate Tampering Detection (Track 2)},
  author       = {{Alibaba Cloud Tianchi}},
  year         = {2024},
  howpublished = {\url{https://tianchi.aliyun.com/competition/entrance/532267}},
  note         = {Tianchi Competition Dataset}
}

@article{wang2026tranxadapter,
      title={TranX-Adapter: Bridging Artifacts and Semantics within MLLMs for Robust AI-generated Image Detection}, 
      author={Wenbin Wang and Yuge Huang and Jianqing Xu and Yue Yu and Jiangtao Yan and Shouhong Ding and Pan Zhou and Yong Luo},
      year={2026},
      eprint={2602.21716},
      archivePrefix={arXiv},
      primaryClass={cs.CV},
      url={https://arxiv.org/abs/2602.21716}, 
}

@misc{openai_gpt_image_2_announcement,
  title        = {Introducing gpt-image-2 -- Available Today in the API and Codex},
  author       = {{OpenAI}},
  year         = {2026},
  howpublished = {\url{https://community.openai.com/t/introducing-gpt-image-2-available-today-in-the-api-and-codex/1379479}},
  note         = {Accessed: 2026-05-13}
}

@inproceedings{qu2023towards,
  title={Towards robust tampered text detection in document image: New dataset and new solution},
  author={Qu, Chenfan and Liu, Chongyu and Liu, Yuliang and Chen, Xinhong and Peng, Dezhi and Guo, Fengjun and Jin, Lianwen},
  booktitle={Proceedings of the IEEE/CVF Conference on Computer Vision and Pattern Recognition},
  pages={5937--5946},
  year={2023}
}

@inproceedings{qu2024towards,
  title={Towards modern image manipulation localization: A large-scale dataset and novel methods},
  author={Qu, Chenfan and Zhong, Yiwu and Liu, Chongyu and Xu, Guitao and Peng, Dezhi and Guo, Fengjun and Jin, Lianwen},
  booktitle={Proceedings of the IEEE/CVF Conference on Computer Vision and Pattern Recognition},
  pages={10781--10790},
  year={2024}
}

@inproceedings{qu2025revisiting,
  title={Revisiting tampered scene text detection in the era of generative AI},
  author={Qu, Chenfan and Zhong, Yiwu and Guo, Fengjun and Jin, Lianwen},
  booktitle={Proceedings of the AAAI Conference on Artificial Intelligence},
  volume={39},
  number={1},
  pages={694--702},
  year={2025}
}

@inproceedings{dancetext,
  title={Detect Any AI-Counterfeited Text Image},
  author={Qu, Chenfan and Zhong, Yiwu and Zhu, Xuekang and Li, Junchi and Jiang, Changjiang and Liu, Jian and Jin, Lianwen},
  booktitle={Proceedings of the IEEE/CVF Conference on Computer Vision and Pattern Recognition},
  year={2026}
}

@inproceedings{dinomac,
  title={DINO-MAC: First-Place Winner Solution of the CVPR2026 Robust DeepFake Detection Challenge},
  author={Qu, Chenfan and  Jin, Lianwen and Li, Junchi and Liu, Jingjing and Yu, Bohan and Xie, Jiangwei and Liu, Jian},
  booktitle={Proceedings of the IEEE/CVF Conference on Computer Vision and Pattern Recognition WorkShops},
  year={2023}
}

@article{qu2025webly,
  title={Webly-Supervised Image Manipulation Localization via Category-Aware Auto-Annotation},
  author={Qu, Chenfan and Zhong, Yiwu and He, Huiguo and Li, Bin and Jin, Lianwen},
  journal={arXiv preprint arXiv:2508.20987},
  year={2025}
}

@article{qu2024textsleuth,
  title={TextSleuth: Towards Explainable Tampered Text Detection},
  author={Qu, Chenfan and Liu, Jian and Chen, Haoxing and Yu, Baihan and Liu, Jingjing and Wang, Weiqiang and Jin, Lianwen},
  journal={arXiv preprint arXiv:2412.14816},
  year={2024}
}

@inproceedings{quomni,
  title={Omni-IML: Towards Unified Interpretable Image Manipulation Localization},
  author={Qu, Chenfan and Zhong, Yiwu and Guo, Fengjun and Jin, Lianwen},
  booktitle={The Fourteenth International Conference on Learning Representations}
}

@inproceedings{qu2026textshield,
  title={TextShield-R1: Reinforced Reasoning for Tampered Text Detection},
  author={Qu, Chenfan and Zhong, Yiwu and Liu, Jian and Zhu, Xuekang and Yu, Bohan and Jin, Lianwen},
  booktitle={Proceedings of the AAAI Conference on Artificial Intelligence},
  volume={40},
  number={10},
  pages={8621--8629},
  year={2026}
}

@misc{du2026SICA,
      title={Can We Build a Monolithic Model for Fake Image Detection? SICA: Semantic-Induced Constrained Adaptation for Unified-Yet-Discriminative Artifact Feature Space Reconstruction}, 
      author={Bo Du and Xiaochen Ma and Xuekang Zhu and Zhe Yang and Chaogun Niu and Chenfan Qu and Mingqi Fang and Zhenming Wang and Jingjing Liu and Jian Liu and Ji-Zhe Zhou},
      year={2026},
      eprint={2602.06676},
      archivePrefix={arXiv},
      primaryClass={cs.CV},
      url={https://arxiv.org/abs/2602.06676}, 
}

@article{schuhmann2022laion,
  title={Laion-5b: An open large-scale dataset for training next generation image-text models},
  author={Schuhmann, Christoph and Beaumont, Romain and Vencu, Richard and Gordon, Cade and Wightman, Ross and Cherti, Mehdi and Coombes, Theo and Katta, Aarush and Mullis, Clayton and Wortsman, Mitchell and others},
  journal={Advances in neural information processing systems},
  volume={35},
  pages={25278--25294},
  year={2022}
}

@inproceedings{tuo2024anytext,
  title={Anytext: Multilingual visual text generation and editing},
  author={Tuo, Yuxiang and Xiang, Wangmeng and He, Jun-Yan and Geng, Yifeng and Xie, Xuansong},
  booktitle={International Conference on Learning Representations},
  volume={2024},
  pages={56783--56799},
  year={2024}
}

@misc{huang2025sofakebenchmarkingexplainingsocial,
      title={So-Fake: Benchmarking and Explaining Social Media Image Forgery Detection}, 
      author={Zhenglin Huang and Tianxiao Li and Xiangtai Li and Haiquan Wen and Yiwei He and Jiangning Zhang and Hao Fei and Xi Yang and Xiaowei Huang and Bei Peng and Guangliang Cheng},
      year={2025},
      eprint={2505.18660},
      archivePrefix={arXiv},
      primaryClass={cs.CV},
      url={https://arxiv.org/abs/2505.18660}, 
}

@misc{sacp,
  title        = "Security AI Challenger Program",
  author       = "{Alibaba Security}",
  howpublished = "\url{https://tianchi.aliyun.com/competition/entrance/531812/introduction}",
  year         = 2020,
}

@article{hu2022lora,
  title={Lora: Low-rank adaptation of large language models.},
  author={Hu, Edward J and Shen, Yelong and Wallis, Phillip and Allen-Zhu, Zeyuan and Li, Yuanzhi and Wang, Shean and Wang, Liang and Chen, Weizhu and others},
  journal={Iclr},
  volume={1},
  number={2},
  pages={3},
  year={2022}
}

\newpage
\beginappendix

\section{Related Works}

\subsection{Datasets}

The evolution of existing FIDL datasets reflects a clear trend from single-forgery detection toward unified multi-domain forensics. Early datasets were typically constructed for specific targets or manipulation types, such as face swapping, natural-image splicing, document text tampering, or full-image generation detection. As a result, they exhibit strong domain-specific properties in data sources, annotation formats, task definitions, and evaluation protocols. With the rapid development of T2I/I2I foundation models, the boundaries between different forgery domains have become increasingly blurred: AIGC datasets now cover real commercial APIs and open-domain generation distributions, document datasets incorporate high-fidelity text-image generation, and Nature/IMDL datasets have expanded from traditional splicing and copy-move operations to generative local editing. Nevertheless, existing datasets still mainly serve their own subdomains and have not yet formed a unified data foundation that can simultaneously support DeepFake, AIGC, Document, and Nature forensics. Therefore, this section reviews four representative categories of datasets and analyzes their limitations for unified FIDL modeling.

\subsubsection{DeepFake}
In the DeepFake domain, dataset evolution has shifted from early, low-quality face swaps to high-fidelity, diverse benchmarks designed to test generalization against unseen manipulations. FaceForensics++ ~\citep{rossler2019faceforensics++} remains a foundational benchmark, offering four distinct manipulation types under varying compression levels, establishing a standard for evaluating robustness. To address the ``overfitting to specific artifacts'' issue, Celeb-DF ~\citep{li2020celeb} introduced a large-scale dataset with significantly improved visual quality and reduced visible artifacts, posing a harder challenge for detectors. The DFDC ~\citep{dolhansky2019deepfake} dataset further expanded the scale and diversity by incorporating thousands of videos from hundreds of actors with varied lighting and poses, simulating real-world social media conditions. More recent efforts like WildDeepFake ~\citep{zi2020wilddeepfake} focus on collecting real-world DeepFake from the internet rather than laboratory-generated ones, providing a critical testbed for in-the-wild performance. Additionally, datasets like UADFV ~\citep{li2018uadfv} and Fsh ~\citep{li2019faceshifter} target specific manipulation techniques, such as inconsistent head poses or occlusion-aware swapping, allowing for fine-grained analysis of detector failures. These collections collectively drive the field towards models that can detect subtle blending boundaries and frequency anomalies across diverse generative backbones.

\subsubsection{AIGC}

The emergence of diffusion models and large-scale generative AI has necessitated a new category of datasets focused on entirely synthetic images rather than manipulated faces. Unlike DeepFake, AIGC datasets evaluate the detection of global texture inconsistencies and latent space artifacts. GenImage ~\citep{zhu2023genimage} stands as a million-scale benchmark, pairing real images from ImageNet with AI-generated counterparts across eight different models, enabling comprehensive evaluation of cross-model generalization. ForenSynths ~\citep{cnnspot} was an early pioneer, aggregating images from multiple GAN architectures to train universal detectors. As diffusion models gained prominence, DiffusionForensics~\citep{dire} and Chameleon ~\citep{aide} emerged to specifically target the unique noise patterns and spectral signatures left by diffusion processes. Datasets like SynthWildx and BFree-Online push the boundary further by including images generated by black-box commercial APIs, which are often post-processed and harder to distinguish from natural photos. Furthermore, FakeInversion ~\citep{fakeinversion} leverages the inversion capabilities of generative models to create hard negative samples, while EvalGEN ~\citep{evalgen} provides a structured framework for evaluating detection reliability across diverse prompts and styles. These datasets are critical for shifting the forensic focus from local facial artifacts to global statistical deviations.

\subsubsection{Document}

Document forgery detection constitutes a specialized subfield with the focus on textual consistency, font rendering, and layout integrity~\cite{qu2024textsleuth}. This domain requires datasets that capture subtle tampering such as character substitution, copy-paste operations, and signature forgery. DocTamper ~\citep{qu2023towards} is a representative large-scale dataset that includes various tampering types on scanned documents, providing pixel-level masks for localization tasks. Building on OCR benchmarks, T-SROIE ~\citep{wang2022tsroie} and Tampered IC13~\citep{wang2022tpic} adapt receipt and scene text datasets by introducing synthetic manipulations to evaluate the robustness of text-based forensic models.  Datasets like RTM~\citep{luo2025rtm} and RIFLC~\citep{rific} aim to bridge the gap between synthetic tampering and real-world scanned artifacts. More recent contributions like DanceText ~\citep{dancetext} and OSTF ~\citep{qu2025revisiting} explore dynamic text manipulations and font inconsistencies, which are common in forged contracts or official certificates. Unlike general image forensics, these datasets emphasize the semantic coherence of text and the physical properties of ink and paper, making them essential for developing models capable of verifying the authenticity of official records and legal documents.

\subsubsection{Nature}

The Nature category encompasses the traditional core of image forensics, focusing on splicing, copy-move, and removal operations in natural scenes such as landscapes, objects, and crowds. This domain relies on datasets that challenge detectors with complex backgrounds, varying lighting conditions, and heavy post-processing. CASIA ~\citep{CASIA_2013} is a seminal dataset that has served as a standard baseline for two decades, covering both splicing and copy-move forgeries with ground-truth masks. To address the limitations of older datasets, IMD2020 ~\citep{IMD20_2020} introduced a large-scale, high-resolution collection with diverse content and realistic tampering, significantly raising the bar for detection performance. Columbia~\citep{Columbia_2006} provides early but rigorous benchmarks focusing on specific artifacts like lighting inconsistencies and double JPEG compression. For copy-move specifically, Coverage~\citep{Coverage_2016} and CocoGlide~\citep{trufor2023} utilize images from the COCO dataset ~\citep{mscoco_2014} to create geometrically transformed duplicates, testing the detector's ability to handle rotation and scaling. Autosplice ~\citep{jia2023autosplice} further expands the diversity by automating the splicing process to generate vast amounts of training data with realistic semantic combinations. Finally, DEFACTO-12k~\citep{defacto_2019} offers a comprehensive suite of post-processing attacks to evaluate robustness. These datasets collectively ensure that forensic models can generalize beyond face-specific cues to detect manipulations in any natural context. 

\subsubsection{Conclusion}
Overall, existing FIDL datasets have advanced detection and localization techniques within their respective subdomains. DeepFake datasets have gradually expanded to high-fidelity, in-the-wild, and large-scale face forgeries; AIGC datasets have evolved from GAN-image detection to diffusion models, commercial APIs, and open-domain generation; Document datasets increasingly emphasize high-fidelity text generation, layout consistency, and realistic scanning noise; and Nature datasets have extended from traditional splicing, copy-move, and removal operations to generative local editing. However, these datasets remain highly fragmented, with substantial differences in data scale, real/fake ratio, supervision granularity, and evaluation objectives, making them insufficient for directly training and evaluating a unified FIDL foundation model. Although ForensicHub~\citep{forensichub} and OpenMMsec~\citep{du2026SICA} have explored unified cross-domain training and evaluation and further revealed potential task conflicts in multi-domain joint training, existing efforts are still largely limited to academic benchmark scales, with restricted data sources, complexity, and domain coverage. In contrast, DefakerOne is designed for unified modeling by systematically integrating tens of millions of heterogeneous samples from AIGC, DeepFake, Document, and Nature domains. Through multi-version data composition experiments, it analyzes the complementarity, transferability, and interference among different artifacts, thereby exploring how fragmented data can be transformed into an effective data mixture and foundation-model capability for full-domain FIDL.




\subsection{Methods}


Existing FIDL methods have evolved from domain-specific vision models to multimodal large language models (MLLMs). Early approaches typically designed specialized detectors for particular forgery types, such as pixel-level localization networks for natural-image manipulation, face forgery classifiers for DeepFake detection, generated-image detectors for AIGC, and text-region localization models for document tampering. While these methods have achieved substantial progress within their own subdomains, their architectures, input-output formats, and artifact assumptions are often highly task-dependent. Recently, MLLMs have been introduced into fake image detection and explanation due to their strong visual-semantic understanding, natural-language interaction, and unified input-output interface, offering new possibilities for unified FIDL modeling. Therefore, this section first reviews traditional vision-based methods across different FIDL subdomains, then discusses recent MLLM-based explorations in unified detection, explanation, and localization, and finally analyzes the remaining gaps toward a unified FIDL foundation model.

\subsubsection{Vision Models for FIDL}
In early fake image forensics, different tasks were long studied as independent subfields, mainly including DeepFake Detection, Image Manipulation Detection and Localization (IMDL), AI-Generated Image Detection (AIGC Detection), and Document Image Manipulation Localization (Doc). These domains differ significantly in their target forgery objects and application scenarios: DeepFake Detection mainly focuses on face-centric manipulations such as face swapping, identity replacement, and expression editing; IMDL targets local tampering in natural images, such as splicing, copy-move, object removal, and inpainting; AIGC Detection aims to identify fully synthetic images generated by generative models; and Doc focuses on localizing tampered text regions in receipts, certificates, contracts, and scene-text images. Therefore, early vision-based forensic methods were not proposed under a unified Fake Image Detection and Localization (FIDL) framework, but instead developed independently around domain-specific forgery types, artifacts priors, output formats, and evaluation protocols.

In the Nature domain~\citep{ma2024imdl}, methods are typically required to perform both image-level detection and pixel-level manipulation localization. MVSS-Net~\citep{mvsspp_2022} adopts multi-view and multi-scale supervision, enhancing manipulation feature learning with noise residuals and boundary inconsistencies. CAT-Net~\citep{CAT-Net2022} focuses on DCT-domain artifacts left by JPEG compression and models compression traces for image manipulation detection and localization. PSCC-Net employs a progressive spatio-channel correlation structure to predict manipulation masks in a coarse-to-fine manner. APSC-Net~\citep{qu2025webly} is designed for modern real-world image manipulation localization, further improving the localization capability for complex tampered regions. IML-ViT~\citep{ma2023iml} introduces Vision Transformers into manipulation localization, using self-attention to model long-range relationships and non-semantic discrepancies among image regions. NCL~\citep{NCL_IML_2023} introduces non-mutually exclusive contrastive learning to exploit contour patches for robust manipulation localization without pre-training. TruFor~\citep{trufor2023} combines RGB content features with learned noise-sensitive fingerprints, treating manipulated regions as anomalies with respect to the overall image consistency. SparseViT~\citep{sparseViT_2025} attempts to move beyond traditional handcrafted feature extractors by using sparse visual modeling to encourage the model to focus more on non-semantic manipulation traces. Mesorch~\citep{zhu2025mesorch} adopts a hybrid CNN-Transformer architecture to model tampered regions at the mesoscopic level between microscopic manipulation traces and macroscopic semantic structures. RITA~\citep{zhu2025does} addresses the dimensional collapse issue in composite image manipulations by reformulating repeated tampering as a temporally ordered process and introducing an autoregressive multi-step localization paradigm to model hierarchical state transitions across manipulation stages.

In the DeepFake Detection domain, the task is usually formulated as image-level binary classification, with a focus on identity, texture, frequency, and blending-boundary anomalies introduced during face generation or editing. Capsule-Net~\citep{nguyen2019capsule} is among the early methods that introduce capsule networks into face forgery detection, using dynamic routing to model spatial relationships among local facial components. RECCE~\citep{cao2022recce} adopts reconstruction-classification learning and exploits reconstruction discrepancies between real and forged faces to improve forgery detection. SPSL~\citep{liu2021spsl} focuses on frequency anomalies in the phase spectrum and captures upsampling artifacts in face forgery through shallow spectral features. SBI~\citep{shiohara2022sbi} constructs self-blended images from a single real face to simulate blending boundaries and source-target statistical inconsistencies commonly observed in DeepFake, thereby improving generalization to unseen forgery methods. Sia~\citep{sia} introduces self-information attention from an information-theoretic perspective, encouraging the model to focus on more discriminative forged regions and channels. Effort~\citep{effort} further decomposes forgery-related and content-related features via orthogonal subspace decomposition, mitigating overfitting to limited forgery patterns.

In the AIGC Detection domain, the goal is mainly to distinguish real images from images synthesized by GANs, diffusion models, or other generative models. DualNet~\citep{dualnet} adopts a dual-stream architecture consisting of a residual stream and a content stream, separately modeling texture residuals and low-frequency content anomalies, and then fusing these cues through cross-attention. HiFiNet~\citep{HiFi-Net2023} proposes a hierarchical fine-grained forgery attribute modeling framework, where a multi-branch feature extractor jointly learns image-level detection features and pixel-level localization-related cues. UnivFD~\citep{univfd} leverages the general feature space of large-scale vision-language pretrained models such as CLIP and performs universal fake image detection across generative models with a simple classifier. FatFormer~\citep{fatformer} introduces a forgery-aware adaptive Transformer, enabling the model to adaptively capture artifacts left by different generative models. Moreover, Forensic-MOE~\citep{forensic-moe} proposes a mix-of-expert architecture to explore comprehensive forensic traces. CO-SPY~\citep{co-spy} combines semantic features and pixel-level artifacts, improving the generalization ability for synthetic image detection through multi-cue fusion. 

In the Doc domain, methods focus on localizing fine-grained and low-visibility tampered text regions in document or scene-text images.  DTD~\citep{qu2023towards} addresses the weak visual differences in document tampering by introducing a Frequency Perception Head and a Multi-view Iterative Decoder, leveraging JPEG frequency compression features and multi-view information to localize tampered text. FFDN~\citep{chen2024ffdn} employs a Visual Enhancement Module and a Wavelet-like Frequency Enhancement module to fuse and decompose frequency features, thereby capturing subtle anomalies in compression traces, textures, and text boundaries. CAFTB~\citep{song2025caftb} adopts a cross-attention-based two-branch design, modeling document tampering traces from both the spatial domain and the noise domain, and fusing the two types of information through cross-attention. TIFDM~\citep{dong2024tifdm} performs forgery traces enhancement and multiscale attention-based localization, enhancing multi-domain tampering traces and combining them with multiscale context for text tampering localization.

Overall, the above vision-based forensic models have achieved significant progress in their respective domains, including DeepFake Detection, IMDL, AIGC Detection, and Doc. However, most of them are still designed for specific tasks, relying on domain-specific data formats, artifacts before training protocols, and output forms. 

\subsubsection{MLLMs for FIDL}

By generating natural language outputs, MLLMs inherently provide human-readable interpretability, offering a distinct advantage over traditional vision models when serving as AIGC or tampered image detectors. Pioneering this direction, AntiFakePrompt~\citep{chang2023antifakeprompt} successfully introduced Vision-Language Models (VLMs) into the FID domain.

Subsequent research has largely focused on balancing detection accuracy, reasoning depth, and inference latency. X2-DFD~\citep{x2-dfd} proposes an explainable and extendable framework based on MLLMs for DeepFake detection. FakeScope~\citep{fakescope}, ThinkFake~\citep{huang2025thinkfake} and Ivy-xDetector~\citep{zhang2025ivyfakeunifiedexplainableframework} employ a CoT paradigm that generates explanations prior to the final prediction, though this necessitates generating a massive number of tokens. To alleviate this latency bottleneck, FakeVLM~\citep{fakeclue} proposed a ``classify-then-explain'' pipeline. While this resolves the latency issue, its natural language explanations remain overly vague and lack specificity; furthermore, extensive experiments reveal its vulnerability in detecting certain tampered images. Addressing the CoT computational overhead, Fake-HR1~\citep{fakehr1} proposed a hybrid reasoning chain specifically designed to reduce the burden in FID scenarios. Similarly, Mirage-R1~\citep{xia2025mirage} and Forensic-Chat~\citep{forensicchat} adopted a ``reason-then-detect'' schema to better guide model perception. 

Recent advancements have expanded MLLMs into specialized domains and advanced learning paradigms. For tampered text images, TextSleuth~\citep{qu2024textsleuth} achieved interpretable detection by incorporating a perception head~\citep{qu2023towards}. Omni-IML~\citep{quomni} mitigated hallucinations through a decoupled design, while TextShield-R1~\citep{qu2026textshield} further enhanced generalization via cross-domain pretraining and reinforcement learning. To achieve semantic-level forgery localization, ForgeryGPT~\citep{liu2024forgerygpt} and FakeShield~\citep{xu2024fakeshield} innovatively integrated a segmentation module, combining the interpretability of MLLMs with the precision of segmentation networks. Other works have explored agent-based frameworks: UniShield~\citep{huang2025unishield} dynamically invokes a comprehensive library of anti-forgery tools based on the specific forgery type to unify the FIDL process, while AgentFoX~\citep{agentfox} utilizes agentic workflows to bolster MLLM interpretability on fake images. In terms of training paradigms, Veritas~\citep{veritas} introduced an innovative reinforcement learning approach, adopting an R1-like multi-stage training process to enhance forgery interpretability. Within DeepFake detection tasks, M2F2-Det~\citep{guo2025rethinking} incorporates pre-trained CLIP features to significantly improve detection performance. Compared with prior reasoning-based methods that primarily focus on global prediction, Legion~\citep{kang2025legion} emphasizes fine-grained artifact grounding, while SIDA~\citep{sida} targets deployment in real-world social media environments. EvoGuard~\citep{zhu2026evoguard} explores a capability-aware dynamic orchestration mechanism for FID. AlignGemini~\citep{chen2025task} achieves generalizable AI-generated image detection through a task-model alignment approach, utilizing a dual-branch architecture that combines a Vision Language Model for high-level semantic consistency checking with a conventional vision model for low-level pixel artifacts detection. Similarly, TranX-Adapter~\citep{wang2026tranxadapter} enhances AIGI detection by injecting texture-level artifact features into MLLMs and designing a lightweight adapter to fuse semantic and artifact representations through optimal-transport-based fusion and cross-attention.

\subsubsection{Conclusion}
In summary, existing FIDL methods have evolved from domain-specific vision detectors to MLLM-driven explainable frameworks, yet a strong foundation-model baseline for full-domain FIDL is still missing. Traditional models are usually tailored to individual subdomains such as DeepFake, IMDL, AIGC, or Document forensics, relying on domain-specific artifacts, specialized architectures, and fixed output formats. As a result, they struggle to handle cross-domain forgery distributions and unified detection-localization requirements. Recent MLLM-based methods provide unified interfaces, visual-semantic understanding, and natural-language explanation ability, but their focus often shifts toward reasoning-chain generation or descriptive interpretation rather than the core FIDL objectives of authenticity detection and forgery localization. This makes it difficult for them to fully exploit large-scale image-level labels and pixel-level masks, or to serve as stable full-domain FIDL baselines. Although ForensicHub~\citep{forensichub} and SICA~\citep{du2026SICA} reveal the trend and challenges of unified FIDL from the perspectives of evaluation protocols and multi-domain training conflicts, they mainly formulate the problem and analyze the difficulties, rather than providing a strong model natively designed for full-domain FIDL. In contrast, DefakerOne returns to the core FIDL tasks by using the visual-semantic capability of MLLMs for authenticity judgment and forgery localization, instead of emphasizing complex explanation generation. Built upon an InternVL2 + SAM2 framework and supported by multi-domain data composition experiments, DefakerOne systematically explores how heterogeneous forensic data can be transformed into unified detection, localization, and generalization capabilities, providing a direct and strong baseline for full-domain FIDL foundation models.
\section{Training Data Composition}
\label{app:data_composition}

Table~\ref{tab:training_data_composition} summarizes the domain-wise composition of the 12.5M training samples used for DeFakerOne. The training data covers four FIDL domains, including DeepFake, AIGC, Document, and Nature. For each domain, we combine public datasets with private real-world data when available, and apply domain-specific sampling to avoid over-dominance from any single dataset.
\begin{table*}[t]
\centering
\resizebox{0.95\linewidth}{!}{
\begin{tabular}{llcccc}
\toprule
Domain & Total & Dataset/Size & Dataset/Size & Dataset/Size & Dataset/Size \\
\midrule

\multirow{3}{*}{DeepFake} 
& \multirow{3}{*}{3.1M}
& \makecell{FF++~\citep{rossler2019faceforensics++}\\0.11M}
& \makecell{CelebDF-v2~\citep{li2020celeb}\\0.18M}
& \makecell{DFD~\cite{dfd2019}\\0.10M}
& \makecell{DFDC~\cite{dfd2019}\\0.09M} \\
& 
& \makecell{ScaleDF~\citep{scaledf}{}\\0.14M}
& \makecell{DF40~\citep{yan2024df40}\\0.10M}
& \makecell{WDF~\citep{zi2020wilddeepfake}\\0.10M}
& \makecell{MFFI~\citep{miao2025mffi}\\0.06M} \\
& 
& \makecell{Private Dataset\\2.22M}
& --
& --
& -- \\
\midrule

\multirow{2}{*}{AIGC}
& \multirow{2}{*}{3.6M}
& \makecell{DiffusionForensics~\citep{dire}\\0.034M}
& \makecell{CommunityForensics~\citep{CommunityForensics}\\0.95M}
& \makecell{GenImage~\citep{zhu2023genimage}\\0.09M}
& \makecell{LAION\_DATA~\citep{schuhmann2022laion}\\1.20M} \\
&
& \makecell{ForenSynths~\citep{cnnspot}\\0.07M}
& \makecell{Private Dataset\\1.256M}
& --
& -- \\
\midrule

\multirow{2}{*}{Document}
& \multirow{2}{*}{2.5M}
& \makecell{DocTamper~\citep{qu2023towards}\\0.145M}
& \makecell{T-SROIE~\citep{wang2022tsroie}\\0.0006M}
& \makecell{RTM~\citep{luo2025rtm}\\0.009M}
& \makecell{SACP~\citep{sacp}\\0.003M} \\
&
& \makecell{RIFLC~\citep{rific}\\0.002M}
& \makecell{OSTF~\citep{qu2025revisiting}\\0.0022M}
& \makecell{Private Dataset\\2.338M}
& -- \\
\midrule

\multirow{2}{*}{Nature}
& \multirow{2}{*}{3.3M}
& \makecell{MIML~\citep{qu2024towards}\\0.937M}
& \makecell{CASIA-v2~\citep{CASIA_2013}\\0.06M}
& \makecell{COCO\_2017~\citep{mscoco_2014}\\0.82M}
& \makecell{OpenSDI~\citep{opensdi}\\0.20M} \\
&
& \makecell{So-Fake-OOD~\citep{huang2025sofakebenchmarkingexplainingsocial}\\0.062M}
& \makecell{So-Fake-Set~\citep{huang2025sofakebenchmarkingexplainingsocial}\\1.20M}
& --
& -- \\

\bottomrule
\end{tabular}
}
\caption{Composition of the DeFakerOne training data.}
\label{tab:training_data_composition}
\end{table*}

\section{Contributors}
\label{sec:contri}

\large{All contributors are listed \textbf{alphabetically by the first name}.} 

\large{
\begin{minipage}[t]{0.48\textwidth} 
    \textbf{Core Contributors} \\
    Changjiang Jiang \\
    Chenfan Qu \\
    Jiangwei Xie \\
    Mingqi Fang \\
    Song Zhou \\
    Xuekang Zhu \\
    \vspace{0.4cm} \\
    \textbf{Contributors} \\
    Chenfeng Zhang \\
    Longfei Liu
\end{minipage}
\hfill
\begin{minipage}[t]{0.48\textwidth} 
    \textbf{Data Leader} \\
    Zhenming Wang \\
    \vspace{0.4cm} \\
    \textbf{Project Leader} \\
    Jian Liu \\
    Jingjing Liu \\
    \vspace{0.4cm} \\
    \textbf{Project Advisor} \\
    Weiqiang Wang
\end{minipage}
}

\begin{table}[!t]
  \caption{Robust analysis (Accuracy) comparison with baseline methods on OpenMMsec~\citep{du2026SICA}.}
  \label{tab:3_robust_full}
  \centering
  \resizebox{0.95\linewidth}{!}{%
  \begin{tabular}{ll|ccccccc}
    \toprule
    \multicolumn{2}{c}{\textbf{Method}} 
    & \textbf{FFDN} 
    & \textbf{Mesorch} 
    & \textbf{Effort} 
    & \textbf{ForensicsAdapter} 
    & \textbf{ForensicMOE} 
    & \textbf{FakeShield} 
    & \textbf{\modelbase} \\
    \midrule
    \multirow{6}{*}{Gaussian Blur}
    & 0.5 & 42.84 & 51.78 & 59.30 & 57.65 & 55.70 & 62.86 & \textbf{85.44} \\
    & 1.0 & 46.32 & 50.43 & 58.43 & 56.70 & 55.13 & 63.46 & \textbf{80.72} \\
    & 1.5 & 48.68 & 49.07 & 56.55 & 55.75 & 53.63 & 65.57 & \textbf{77.59} \\
    & 2.0 & 49.68 & 48.67 & 55.23 & 54.95 & 53.13 & 65.54 & \textbf{76.81} \\
    & 2.5 & 50.62 & 47.30 & 54.03 & 53.77 & 52.95 & 65.74 & \textbf{76.74} \\
    & Avg & 47.63 & 49.45 & 56.71 & 55.76 & 54.11 & 64.63 & \textbf{79.46} \\
    \midrule
    \multirow{6}{*}{Brightness}
    & 0.5 & 44.65 & 53.00 & 57.88 & 56.17 & 55.35 & 64.01 & \textbf{81.44} \\
    & 1.0 & 39.68 & 51.58 & 58.90 & 57.75 & 55.70 & 63.14 & \textbf{84.59} \\
    & 1.5 & 44.59 & 50.80 & 58.75 & 57.15 & 55.33 & 64.34 & 50.33 \\
    & 2.0 & 46.78 & 50.03 & 57.15 & 55.67 & 54.15 & 63.61 & \textbf{70.36} \\
    & 2.5 & 48.22 & 49.87 & 54.95 & 53.80 & 53.58 & 62.31 & \textbf{66.26} \\
    & Avg & 44.78 & 51.06 & 57.53 & 56.10 & 54.82 & 63.48 & \textbf{70.60} \\
    \midrule
    \multirow{6}{*}{Contrast}
    & 0.5 & 44.61 & 52.28 & 58.58 & 56.00 & 55.03 & 63.41 & \textbf{80.97} \\
    & 1.0 & 39.68 & 51.58 & 58.90 & 57.75 & 55.70 & 64.69 & \textbf{84.52} \\
    & 1.5 & 43.38 & 50.80 & 59.05 & 56.70 & 56.10 & 64.31 & \textbf{76.24} \\
    & 2.0 & 44.88 & 50.45 & 57.40 & 55.97 & 55.50 & 64.31 & \textbf{71.06} \\
    & 2.5 & 45.68 & 50.50 & 57.43 & 54.72 & 55.30 & 62.76 & \textbf{68.51} \\
    & Avg & 43.65 & 51.12 & 58.27 & 56.22 & 55.53 & 63.86 & \textbf{76.26} \\
    \midrule
    \multirow{6}{*}{JPEG Compression}
    & 75 & 39.95 & 51.78 & 58.68 & 56.67 & 51.35 & 62.94 & \textbf{78.36} \\
    & 80 & 39.58 & 52.30 & 58.75 & 56.95 & 50.88 & 62.69 & \textbf{76.59} \\
    & 85 & 39.38 & 52.00 & 57.13 & 56.37 & 51.25 & 63.46 & \textbf{75.29} \\
    & 90 & 40.75 & 51.63 & 59.58 & 56.80 & 51.55 & 62.01 & \textbf{75.04} \\
    & 95 & 40.01 & 52.25 & 59.53 & 57.15 & 53.33 & 62.74 & \textbf{75.51} \\
    & Avg & 39.93 & 52.00 & 58.73 & 56.78 & 51.67 & 62.77 & \textbf{76.16} \\
    \midrule
    \multirow{6}{*}{Noise}
    & 0.05 & 45.28 & 47.82 & 57.28 & 55.20 & 53.68 & 63.66 & \textbf{71.66} \\
    & 0.1 & 47.88 & 47.42 & 55.53 & 52.52 & 55.95 & 63.66 & \textbf{64.83} \\
    & 0.15 & 47.58 & 47.37 & 54.58 & 52.50 & 54.10 & 63.26 & \textbf{61.23} \\
    & 0.2 & 48.98 & 48.27 & 53.30 & 52.15 & 52.70 & 63.54 & 63.36 \\
    & 0.25 & 50.42 & 50.00 & 52.50 & 51.50 & 51.68 & 63.59 & \textbf{65.53} \\
    & Avg & 48.03 & 48.18 & 54.62 & 52.77 & 53.62 & 63.54 & \textbf{65.32} \\
    \midrule
    \multirow{6}{*}{Resize}
    & 128 & 49.48 & 50.53 & 57.18 & 54.02 & 52.50 & 52.13 & \textbf{61.21} \\
    & 256 & 45.32 & 49.32 & 58.43 & 57.25 & 54.15 & 50.85 & \textbf{69.21} \\
    & 384 & 43.18 & 50.18 & 59.55 & 58.20 & 53.90 & 50.73 & \textbf{71.14} \\
    & 512 & 40.51 & 50.98 & 59.55 & 58.17 & 53.68 & 51.35 & \textbf{70.29} \\
    & 640 & 42.48 & 51.70 & 58.75 & 57.50 & 53.10 & 51.43 & \textbf{74.32} \\
    & Avg & 44.19 & 50.54 & 58.69 & 57.02 & 53.47 & 51.30 & \textbf{69.23} \\
    \midrule
    \multirow{6}{*}{Saturation}
    & 0.5 & 40.75 & 52.38 & 58.93 & 57.25 & 54.95 & 64.69 & \textbf{82.24} \\
    & 1.0 & 39.68 & 51.58 & 58.90 & 57.75 & 55.70 & 63.74 & \textbf{84.57} \\
    & 1.5 & 39.95 & 51.65 & 58.50 & 57.05 & 56.10 & 64.69 & \textbf{83.17} \\
    & 2.0 & 40.95 & 51.23 & 58.08 & 57.37 & 55.58 & 64.56 & \textbf{80.79} \\
    & 2.5 & 42.11 & 51.60 & 57.73 & 57.10 & 56.53 & 64.01 & \textbf{78.49} \\
    & Avg & 40.69 & 51.69 & 58.43 & 57.30 & 55.77 & 64.30 & \textbf{81.85} \\
    \bottomrule
  \end{tabular}
  }
\end{table}






\end{document}